%% file: main_withSuveryAnalysis.tex
\documentclass[10pt,twocolumn]{article}

\usepackage[utf8]{inputenc}
\usepackage[a4paper,margin=1.5cm]{geometry} % <-- control margins
\usepackage{lipsum} % dummy text, remove later
\usepackage{authblk} % affiliations
\usepackage{graphicx} % figures
\usepackage{amsmath, amssymb}
\usepackage{hyperref}
\usepackage{caption}
\usepackage{pdfpages}
\usepackage{booktabs}
\usepackage{array}

\usepackage{booktabs}
\usepackage{tabularx} % Allows for column wrapping
\usepackage{amsmath}  % For better math handling
\usepackage{comment}
\usepackage{ulem}
\usepackage{multirow}
\usepackage{colortbl}
\usepackage{microtype}

\title{Perception-Aligned AI Outputs: End-to-End Visual Prediction for Uncertainty Communication in Clinical Decision-Making}

\author[1,2,*]{Mohammad Eslami} %mohammad_eslami@meei.harvard.edu
\author[2]{Solale Tabarestani} %solale.tabarestani@gmail.com
\author[1]{Saber Kazeminasab} %skazeminasab1992@gmail.com
\author[3]{Ehsan Adeli}%eadeli@stanford.edu
\author[4]{Glyn Elwyn}%glynelwyn@gmail.com
\author[1]{Tobias Elze}%tobias_elze@meei.harvarad.edu
\author[1]{Mengyu Wang}%mengyu_wang@meei.harvard.edu
\author[5]{Nazlee Zebardast}%Nazlee_zebardast@meei.harvard.edu
\author[5]{Lucia Sobrin}%lucia_sobrin@meei.harvard.edu
\author[6]{Nassir Navab}%nassir.navab@tum.de
\author[7,8]{Daniel Shu Wei Ting}%daniel.ting45@gmail.com
\author[2,*]{Malek Adjouadi} %adjouadi@fiu.edu

\affil[1]{Harvard Ophthalmology AI Lab, Schepens Eye Research Institute of Massachusetts Eye and Ear, Harvard Medical School, Boston, MA, USA}
\affil[2]{Center for Advanced Technology and Education, Florida International University, Miami, FL, USA}
\affil[3]{Department of Psychiatry \& Behavioral Sciences, Stanford University, Stanford, CA, USA}
\affil[4]{Dartmouth Institute for Health Policy and Clinical Practice, Dartmouth College, Hanover, NH, USA}
\affil[5]{Massachusetts Eye and Ear, Harvard Medical School, Boston, MA, USA}
\affil[6]{Computer Aided Medical Procedures, Technical University of Munich, Munich, Germany}
\affil[7]{Singapore Eye Research Institute, Singapore National Eye Centre, Singapore, Singapore}
\affil[8]{Department of Ophthalmology, Byers Eye Institute, Stanford University, Palo Alto, CA, USA}
\affil[*]{Corresponding authors: mohammad\_eslami@meei.harvard.edu; adjouadi@fiu.edu}

\date{}

\begin{document}

\maketitle

%%%%%%%%%%%%%%%%%%%
%%%%%%%%%%%%%%%%%%%
%%%%%%%%%%%%%%%%%%%%
%%%%%%%%%%%%%%%%%%%%
%%%%%%%%%%%%%%%%%%%%%

\begin{abstract}
\input{sec_abstract}

\end{abstract}

%%%%%%%%%%%%%%%%%%%
%%%%%%%%%%%%%%%%%%%

% keywords
% Explainable Artificial Intelligence (XAI)
% Human-Centered AI
% Uncertainty Quantification
% Clinical Decision Support
% Medical AI
% Visualization in Machine Learning
% Interpretability
% Healthcare 4.0
% Human-AI Interaction
% Trustworthy AI

%%%%%%%%%%%%%%%%%%%
%%%%%%%%%%%%%%%%%%%
%%%%%%%%%%%%%%%%%%%%
%%%%%%%%%%%%%%%%%%%%
%%%%%%%%%%%%%%%%%%%%%

\section{Introduction}
\input{sec_intro}

% more references: https://academic.oup.com/jamia/issue/31/2

%%%%%%%%%%%%%%%%%%%%%%%%%%%%%%%%%%%
%%%%%%%%%%%%%%%%%%%%%%%%%%%%%%%%%%%%
%%%%%%%%%%%%%%%%%%%%%%%%%%%%%%%%%%%%%%%%%%%%%%%%%%%%%%%%%%%%%%%%%%%%%%%%
%%%%%%%%%%%%%%%%%%%%%%%%%%%%%%%%%%%%%%%%%%%%%%%%%%%%%%%%%%%%%%%%%%%%%%%%
%%%%%%%%%%%%%%%%%%%%%%%%%%%%%%%%%%%%

\section{Methods}

\input{sec_method}

%%%%%%%%%%%%%%%%%%%

\section{Results}
\input{sec_results}

%%%%%%%%%%%%%%%%%%%%%%%%%%%%%%%%%%%%

\section{Survey Analysis}
\input{sec_survey}

%%%%%%%%%%%%%%%%%%%%%%%%%%%%%%%%%%%%
%%%%%%%%%%%%%%%%%%%%%%%%%%%%%%%%%%%%
%%%%%%%%%%%%%%%%%%%%%%%%%%%%%%%%%%%%%%%%%%%%%%%%%%%%%%%%%%%%%%%%%%%%%%%%
%%%%%%%%%%%%%%%%%%%%%%%%%%%%%%%%%%%%%%%%%%%%%%%%%%%%%%%%%%%%%%%%%%%%%%%%
%%%%%%%%%%%%%%%%%%%%%%%%%%%%%%%%%%%%

\section{Discussion}
\input{sec_disscussion}

% =============================================================================
\section{Conclusion}
\input{sec_conclusion}

\section*{Data Availability}
Only publicly available datasets are used for the experiments, and since the characteristics of each experiment are different, the details are mentioned in the corresponding subsections and also appendix.

\section*{Code Availability}
Code is available at: \url{https://github.com/mohaEs/VL4ML}.

\section*{Acknowledgements}
This work was supported by the National Science Foundation (NSF) under Grants CNS-1920182, CNS-1532061, CNS-1338922, CNS-2018611, and CNS-1551221, and NIH (NIA1P50AG047266-01A1) through the Florida Alzheimer’s Disease Research Center.

%\section*{Author Contributions}
%A.B. conceived the study; C.D. performed experiments; E.F. wrote the manuscript.

\bibliographystyle{unsrt}
\bibliography{refs.bib}

\input{appendix}

\end{document}

%% file: sec_abstract.tex
Explainable Artificial Intelligence (XAI) is critical for enabling
trust, transparency, and adoption of AI systems in Healthcare~4.0.
However, conventional XAI methods often rely on static, technical
explanations that are difficult for clinicians and patients to
interpret in practice. In this work, we introduce a new human-centered
explainability paradigm, termed Visualized Learning for Machine
Learning (VL4ML), which directly generates intuitive visual
representations of model outputs to support clinical decision-making
and uncertainty understanding. Unlike traditional post-hoc explanation
techniques, VL4ML produces end-to-end visual outputs in which colors,
patterns, and spatial structures encode diagnostic and prognostic
information, enabling users to interpret predictions without requiring
access to model internals or statistical expertise. Importantly, the
framework enables per-case uncertainty communication in a perceptually
intuitive manner.
 
We evaluate VL4ML across multiple clinical tasks -- classification,
regression, longitudinal prediction, and multimodal analysis --
demonstrating its flexibility across diverse healthcare scenarios.
To assess its effectiveness as an explainability mechanism, we conduct
a large-scale human-centered evaluation ($n = 158$, including 39.2\%
clinical professionals) alongside an expert-based interpretability
study. Positive reception of the visual format exceeded 79\% across
all tasks and evaluation dimensions. Visual outputs were rated as more
memorable than numeric outputs by 84.0\% of participants and as
enabling faster decision-making by 76.9\%. Critically, chi-square
tests with Bonferroni correction found no significant differences in
preference or interpretability between clinicians and non-clinicians,
or between male and female participants, confirming universal
accessibility of the representation. Uncertainty embedded in visual outputs was perceived
by over 82\% of respondents across tasks, without requiring prior
statistical training.
 
These findings highlight VL4ML as an XAI paradigm that bridges the
gap between model predictions and human perception, offering a
complementary and universally accessible perspective to existing
post-hoc explainability and uncertainty quantification approaches,
while supporting more transparent, trustworthy, and user-adaptive
decision-making in healthcare settings.

%% file: sec_intro.tex
Artificial Intelligence (AI) is increasingly integrated into healthcare systems, supporting a wide range of tasks including diagnosis, prognosis, and clinical decision-making. 
Despite significant advances in model performance, the adoption of AI in real-world clinical environments remains limited, largely due to concerns regarding transparency, trust, and interpretability. Clinicians and patients are often required to make critical decisions based on model outputs that are presented as probabilities or abstract explanations, which may not align with human reasoning processes.

Explainable Artificial Intelligence (XAI) has emerged as a key direction to address these challenges \cite{DARPA_XAI, longo2024explainable}. Traditional XAI approaches, such as saliency maps, feature attribution methods, and post-hoc explanation techniques, aim to provide insights into model behavior. However, these methods often remain technical, static, and difficult to interpret for non-expert users. As a result, they may fail to effectively bridge the gap between complex model representations and human understanding, particularly in high-stakes healthcare settings \cite{panayides2025position}. 
In clinical practice, explanations must support rapid decision-making under uncertainty, often by users without technical expertise. Existing XAI methods rarely provide explanations in forms that align with human perception or clinical reasoning workflows \cite{ghanvatkar2024evaluating}.

%In the context of this study, we emphasize key aspects of psychology in XAI: 1) Trust and Confidence: Research in XAI examines how explanations influence users' trust and confidence in AI systems. 2) Comprehensibility: Ensuring that explanations are presented in formats, visualization techniques, and language that are comprehensible to users with varying levels of expertise is a central focus. 3) User Preferences: Recognizing that individuals have diverse preferences for the depth and type of explanations they seek from AI systems, addressing these preferences is essential. 4) Decision Support: In cases where AI systems are designed to assist decision-makers, research explores how explanations can effectively support human decision-making processes.

\begin{figure*}[ht]
\centering
\minipage{\textwidth}
  \includegraphics[width=\linewidth,trim={1cm 4.0cm 1cm 3.5cm},clip]{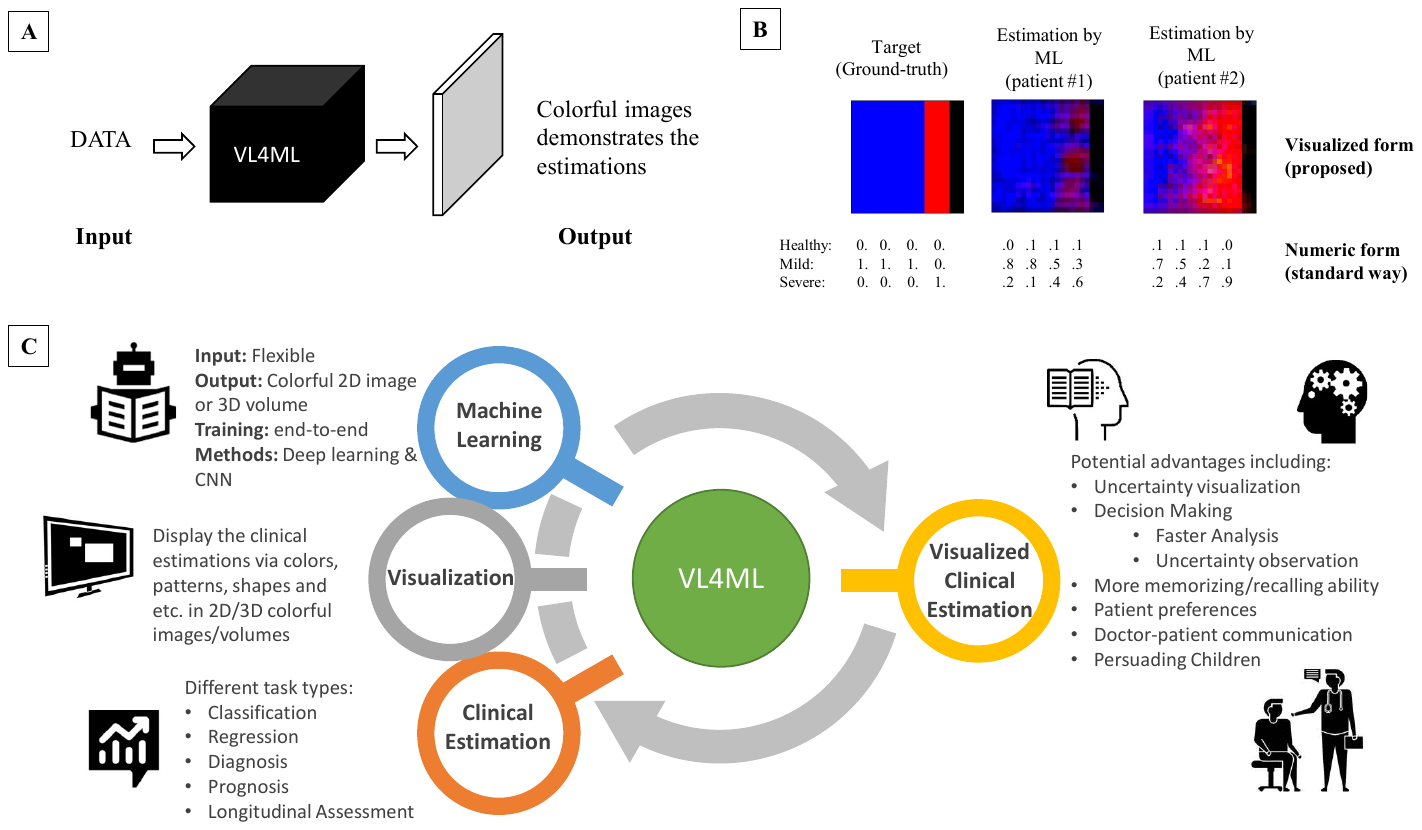}
\endminipage
%\caption{A) The conceptual demonstration of the proposed \textit{VL4ML} approach. The core concept revolves around the generation of an image or visual display where colors, intensities, and patterns convey the diagnostic, prognostic, and predictive outcomes. Importantly, this approach eliminates the need for a two-step pipeline, as the machine directly produces estimations in a visual format. B) Comparing the outputs of the \textit{VL4ML} approach and standard ways with an example of the target outcomes for patients whose disease status is converted from Mild to Severe at the 4th time-point. C) The circumstance of the proposed approach.}
\caption{Overview of the proposed VL4ML paradigm for human-centered explainability and uncertainty communication.
(A) Conceptual illustration of VL4ML, where machine learning models generate visual outputs in which colors, intensities, and spatial patterns encode predictive information along with associated uncertainty.
(B) Comparison between conventional numeric outputs (e.g., predicted probabilities and confidence values) and the proposed visual representations, highlighting how clinically relevant information and uncertainty can be conveyed in an interpretable visual form.
(C) General framework of VL4ML, integrating machine learning, visualization, and clinical estimation to support XAI across multiple tasks, including classification, regression, diagnosis, prognosis, and longitudinal assessment. The approach is intended to complement existing explainability and uncertainty quantification methods and motivate further investigation into perceptually aligned representations.}
\label{fig:intro}
\end{figure*}

%In medical data analysis, machine learning (ML) has shown its effectiveness in producing clinical-grade results comparable to healthcare providers \cite{AIinMed, IntroBook-2019, gerenral-nature-hpc, JAMA-general}. 
%The goal is not a man versus machine competition, but rather, integrating AI as assistive tools in medical decision-making, aligning with healthcare guidelines promoting shared decision-making \cite{gerenral-nature-hpc, JAMA-Shared}. To enhance comprehension, AI algorithms should provide visually intuitive outcomes, overcoming the complexity of data and the opacity of ML \cite{IEEE_MDvisualization, nature_roadmap, theis2016age, valdez2016human}.

%In this study, we introduce Visualized Learning for Machine Learning (VL4ML), which creates machine predictions as intuitive visual displays to improve communication with both patients and professionals. Addressing the critical need for \textit{uncertainty communication} in medical AI \cite{defer3, Nature_UncertaintyQuanti}, VL4ML further incorporates per-case uncertainty visualization. Figures \ref{fig:intro}A and B illustrate this concept.

In this study, we introduce Visualized Learning for Machine Learning (VL4ML), a novel human-centered paradigm that translates model outputs into end-to-end visual representations. Unlike conventional post-hoc methods, VL4ML encodes predictive information and uncertainty directly into colors, patterns, and spatial structures, aligning model interpretation with natural human perception. This approach facilitates faster, more intuitive communication for both patients and professionals. Critically, VL4ML addresses the need for uncertainty communication in medical AI \cite{defer3, Nature_UncertaintyQuanti, ghanvatkar2024evaluating} by incorporating per-case uncertainty visualization (Figures \ref{fig:intro}A and B).

\subsection{Motivation and Advantages of the Generalized Construct}

\uline{XAI 2.0 in Healthcare 4.0:} Recently, there has been a growing recognition of the need for human-centered explainability, often referred to as Explainable AI 2.0 \cite{longo2024explainable}. This paradigm emphasizes explanations that are not only faithful to the model but also aligned with human perception, contextual understanding, and decision-making processes. In this context, explanations should be intuitive, adaptive, and actionable, enabling users to interpret model outputs without requiring detailed knowledge of underlying algorithms. Within the emerging XAI 2.0 paradigm, explainability is no longer limited to post-hoc interpretation, but extends to designing model outputs that are inherently interpretable by humans.

\uline{Psychology of explanation}: On the other side, the psychology of explanation in XAI is a growing multidisciplinary field that examines how people interpret and interact with AI-generated explanations \cite{DARPA_XAI, XAI_psychology, XAI_psychology_2022}. In healthcare, this perspective is especially important, as effective explanations influence trust, confidence, and decision-making. For medical AI to achieve widespread adoption, systems must not only deliver accurate predictions but also embody the principles of trustworthy and responsible AI. 
This requires outputs that are comprehensible to diverse users, align with their preferences, and communicate uncertainty in ways that support clinical reasoning. 

Figure \ref{fig:intro}C summarizes the ecosystem of the proposed approach and its many application domains in support of its generalized construct. Visualization is a powerful tool for data-driven decision-making, enabling trends, patterns, and clusters to be easily recognized \cite{IEEE_MDvisualization, TNNLS-2}. Building on insights from the psychology of explanation, we propose a new paradigm for medical estimation and decision support that leverages visualization for human comprehension, memory, and judgment. There are several aspects that guide the motivation behind this visualization-based estimation approach, among them are: \\
\uline{Faster Brain Analysis}: 
From a human standpoint, visualization works because humans respond to and process visual data better than any other sort of data.
In reality, the human brain analyzes images 60,000 times quicker than words \cite{lindsay2013human-book, potter2014detecting}. We may harness our natural visual abilities to improve data processing and organizational performance. \\
\uline{Enhanced Memorization and Recall Ability}: 
Superior memorization and recall for pictures or images over words is empirically well supported, known as \textit{Picture Superiority Effect} \cite{baadte2019picture, grady1998neural}. 
\\
\uline{Augmented Decision-Making}: 
Visualization enhances decision-making by leveraging the human brain's preference for visual data. It processes images faster than text, accelerating informed decisions \cite{lindsay2013human-book}. Moreover, studies confirm that visualizing data and results boosts confidence and improves decision accuracy \cite{cardoso2016graph, padilla2018decision-review, patient_pref_ieee_review}.
\\
\uline{Patient Preference Supports Visualization}: 
Patient-centered point-of-care systems and shared decision-making strategies strive to provide patients with meaningful visual representations of information \cite{patient_pref_ieee_review, patient_pref_2}. %cite{ patient_pref_4}
Enhanced communication between doctor and patient can also be established with the help of visualization. 
Additionally, the ease of understanding and the visual memorization aid make it more effective and engaging, particularly among pediatric patients \cite{patients_pref_kid_3}. \\
\uline{Visualizing Uncertainty Injected through Machine Learning}:
In decision-making, recognizing uncertainty is crucial. Machine learning, based on statistical analysis and data populations, acknowledges the presence of uncertainty, out-of-distribution samples, and potential bias for each new case or patient. For effective machine-assisted medical decision-making, quantifying uncertainty per patient/case is vital \cite{defer3, Nature_UncertaintyQuanti, bias}. When faced with ambiguity, ML should be capable of abstaining from predictions, seeking human expertise (i.e., 'learning to defer'), or collecting additional data \cite{defer2, defer3, Nature_UncertaintyQuanti}.
Hence, for a transparent presentation of algorithm outcomes to physicians and patients, ML methods must accommodate data- or algorithm-induced uncertainty for each case and be capable of communicating uncertainty. Our proposed approach visualizes some sort of uncertainty, providing a basis for subjective or objective quantification.

\subsection{Contributions}

\iffalse
This study serves as a proof of concept and focuses on generating visual displays of clinical estimations, harnessing the advantages mentioned earlier in a highly practical visualized format. It marks a significant stride toward establishing a comprehensive end-to-end pipeline for visualized learning in decision-making. The key contributions of this study are summarized as follows: 
\hspace{0.05cm} 1) Introducing a conceptual approach using deep learning and convolutional neural networks for visual clinical estimations. \hspace{0.05cm}
2) Uncertainty visualization through deep ensemble technique. \hspace{0.05cm}
3) Demonstrating versatility through four case studies covering different task types. \hspace{0.05cm}
4) Conducting an extensive survey analysis to gather user feedback, affirming the effectiveness of our visualization approach in enhancing decision-making, improving user preferences, and highlighting the benefits of uncertainty visualization. \hspace{0.05cm}
5) Providing open-source code for validation and further research (\url{https://github.com/mohaEs/VL4ML}).
\fi

In this work, we introduce Visualized Learning for Machine Learning (VL4ML), a novel human-centered explainability paradigm that directly generates visual representations of model outputs. Unlike conventional post-hoc explanation methods, VL4ML produces end-to-end visual outputs in which colors, patterns, and spatial structures encode predictive information and uncertainty. This design allows users to interpret model outputs in a manner that is closer to natural human perception, facilitating faster and more intuitive understanding. Consequently, VL4ML defines a new paradigm for XAI 2.0, specifically tailored for the demands of Healthcare 4.0.

The main contributions of this work are as follows: 
\begin{enumerate}
    \item We propose VL4ML, a novel human-centered explainability paradigm that generates interpretable visual outputs directly from machine learning models.
    \item We introduce a new paradigm for explainability that shifts from numerical and textual outputs toward perception-aligned visual representations.
    \item We provide a comprehensive evaluation across multiple clinical tasks, demonstrating the flexibility and applicability of the proposed approach.
    \item We conduct human-centered and expert-based studies showing that the visual outputs preserve clinically meaningful information and support intuitive interpretation.    
\end{enumerate}

By aligning AI outputs with human perception and reasoning, this work contributes to the development of more transparent, trustworthy, and user-adaptive AI systems for Healthcare 4.0.

%% file: sec_method.tex
\subsection{Related Work}

\textit{Post-hoc Explainability Methods:} 
A large body of work in explainable artificial intelligence focuses on post-hoc interpretation of trained models. Feature attribution methods such as SHAP and LIME approximate the contribution of input features to model predictions using local surrogate models or game-theoretic formulations. In the context of deep learning, gradient-based visualization techniques such as Grad-CAM generate saliency maps by highlighting regions that influence the model’s decision.
While these approaches provide insight into model behavior, they are typically limited to highlighting input importance rather than conveying the full predictive structure. In medical imaging, saliency maps are often noisy, sensitive to model perturbations, and difficult to interpret quantitatively. Moreover, these methods require users to interpret abstract heatmaps, which may not directly translate to clinically meaningful decision support \cite{dwivedi2023explainable, molnar2020interpretable}.

\textit{Intrinsic and Interpretable Model Design}:
An alternative direction focuses on designing inherently interpretable models, such as attention-based architectures \cite{bibal2022attention}, prototype learning methods, and concept bottleneck models \cite{koh2020concept}. These approaches aim to embed interpretability within the model structure itself, rather than relying on post-hoc explanations.
However, even intrinsically interpretable models often expose intermediate representations (e.g., attention weights or learned concepts) that remain abstract and require technical expertise to interpret. As a result, their usability in real-world clinical workflows is still limited, particularly for non-technical users.

\textit{Uncertainty Quantification in Medical AI}:
Uncertainty estimation plays a central role in safe and reliable clinical AI systems \cite{salvi2025explainability}. Bayesian approaches such as Monte Carlo Dropout approximate predictive uncertainty by performing stochastic forward passes at inference time. Ensemble-based methods further improve robustness by aggregating predictions from multiple independently trained models. Other approaches, including test-time augmentation and entropy-based measures, have also been widely used to capture epistemic and aleatoric uncertainty.
In parallel, calibration techniques such as temperature scaling and conformal prediction aim to ensure that predicted probabilities reflect true likelihoods. While these methods provide quantitative measures of uncertainty, they typically express uncertainty as scalar values (e.g., variance, entropy, confidence scores), which may be difficult for end-users to interpret in practice \cite{huang2024review, he2026survey, wang2023calibration}. These representations are often not aligned with human perceptual reasoning, limiting their effectiveness for rapid clinical interpretation.

\textit{Visualization of Model Outputs and Predictions}:
Visualization has been used extensively to support the interpretation of machine learning models and clinical data. In medical imaging, visualization techniques often focus on overlaying saliency maps, feature maps, or segmentation outputs on input images. Other work explores dimensionality reduction or trajectory visualization for longitudinal data.
Despite these advances, visualization is commonly applied as a post-processing step, rather than being integrated into the prediction mechanism itself. As a result, the visualization remains auxiliary to the prediction, rather than serving as the primary medium of interpretation.

%%%%%%%%%%%%%%%%%%%%%%%%%%%%%%%%%%%%
\begin{figure*}[tbh]
\centering
\minipage{\textwidth}
  \includegraphics[width=\linewidth,trim={4cm 3.6cm 5.5cm 2cm},clip]{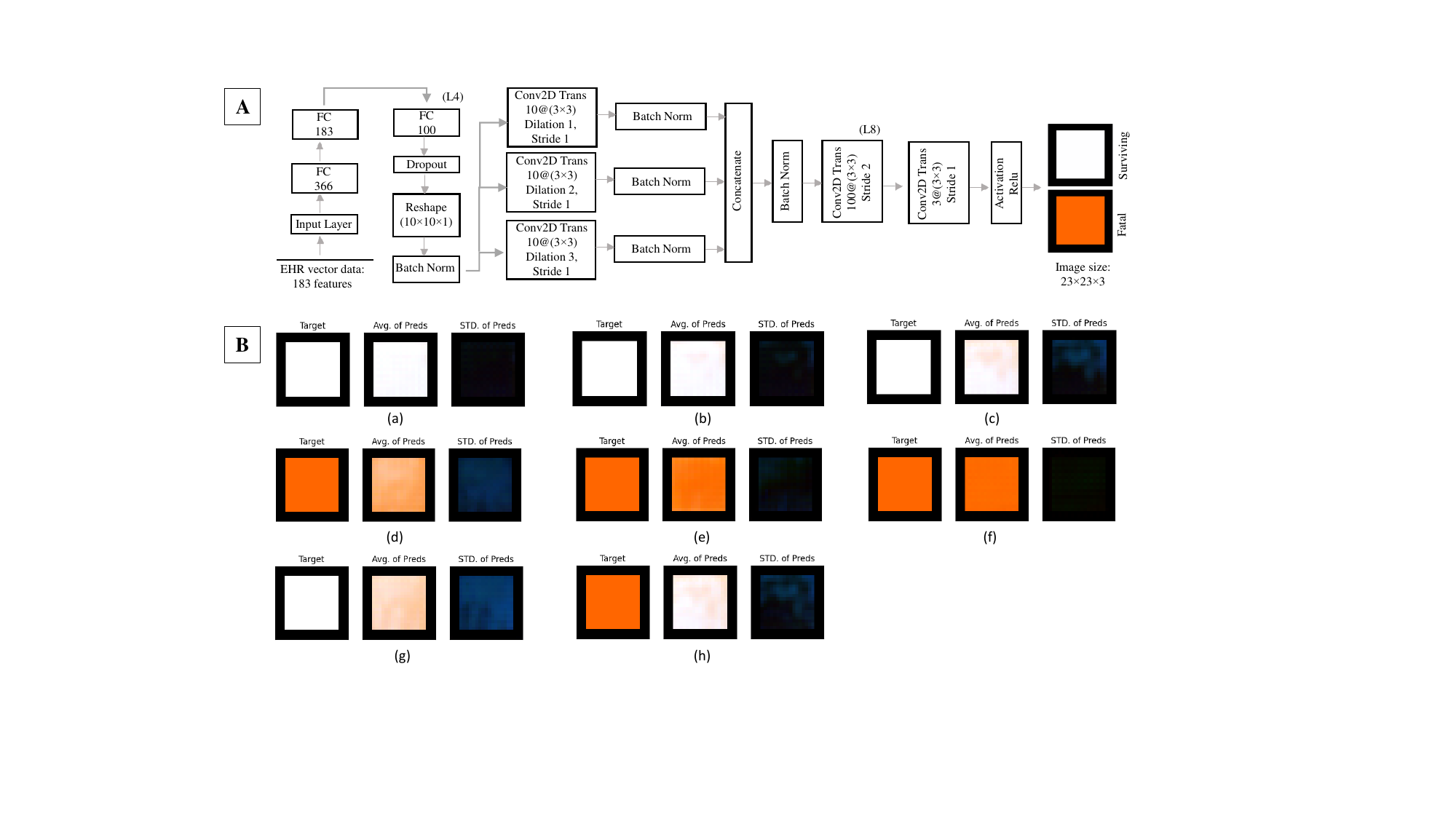}
\endminipage
\caption{A) Network Architecture: This section outlines the architectural details of the network developed for Experiment 1, which focuses on predicting patient survival in the ICU.
B) Experimental Results: The sub-figures present the results for eight distinct test cases involving different patients. Each sub-figure includes three images: the target, the output (average of predictions through deep ensemble, denoted as 'Avg. of Preds'), and the standard deviation of each prediction (referred to as 'STD of Preds').
In this experiment, two types of target images are employed: white for cases indicating survivability (a, b, c, g) and orange for mortality (d, e, f, h), including the surrounding RU area. Notably, the last two sub-figures represent instances of misclassified samples, highlighting the impact of system uncertainty on the generated output.}
\label{fig:survival}
\end{figure*}
%%%%%%%%%%%%%%%%%%%%%%%%%%%%%%%%%%%%

\subsection{Proposed Paradigm}

%Positioning of Our Work

In contrast to prior approaches, VL4ML does not rely on post-hoc explanations, feature attribution, or scalar uncertainty estimates. Instead, it generates end-to-end visual outputs in which predictive information and uncertainty are encoded directly through perceptually meaningful patterns. This shifts visualization from an auxiliary explanation tool to the primary representation of model output. By integrating prediction, uncertainty communication, and interpretation into a unified representation, VL4ML defines a fundamentally different and complementary paradigm for explainability that aligns with the principles of Explainable AI 2.0.
%In contrast to existing approaches, we propose a fundamentally different paradigm in which explainability is embedded directly into the model output. 
%Rather than generating predictions followed by post-hoc explanations, the proposed VL4ML framework produces visual representations as the primary output of the model. 
%These representations encode predictive information and uncertainty in a form that is directly interpretable by humans.

By integrating visualization, prediction, and uncertainty communication into a unified framework, this work aligns with the principles of XAI 2.0 and addresses key limitations of existing explainability methods. The proposed approach shifts the focus from explaining models to designing human-interpretable outputs, enabling more intuitive, transparent, and actionable AI systems for healthcare.

Figure \ref{fig:intro}A shows the concept of the proposed approach through a black box.
The main idea is to produce an image or a visual display wherein colors, intensities, and patterns are to express the resulting diagnosis, prognosis, and prediction outcomes.
The first step is therefore the design of the visual display itself as target or ground-truth images.
Additionally, we propose to have a black area with zero values to display and reserve as the uncertainty area (\textit{RU}), and this criterion could serve as the simplest uncertainty visualization/quantification if the output produced by the machine has affected such a region. \\
Next, we need to develop a deep-learning network capable of producing the results as images. The design of the network is flexible, but we need to embed some layers to produce an output image as well as layers for feature extraction and fusion. 
The output image should be a tensor of size NxNx3, where NxN defines the image size and 3 is to account for the RGB channels.
Obviously, if the input data is not in the form of 2D or 3D tensors, reshaping the features into 2D/3D tensor form becomes necessary. \\
As an example, Figure \ref{fig:survival}A shows the network developed for experiment 1, the survival prediction in the Intensive Care Unit (ICU). \\
Till now, with this trick, we will produce medical estimations in a visualized form instead of a numerical or probability form. Lastly, to make sure we also visualize uncertainty, we use the \underline{deep ensemble} technique \cite{deep_ensembles} and show the average and variance of output images. 

The loss criterion between the target image and the network's output is the mean of absolute error (MAE). The deep ensemble of all the experiments is conducted with 5 models. Datasets are randomly partitioned to train (66\%) and test (33\%). Training is early stopped, monitoring the loss of validation where 10\% of the training set is used as the validation set. More details are available in the appendix.

%% file: sec_results.tex
As a proof of concept, four different experiments were conducted to demonstrate the adaptability of the proposed \textit{VL4ML} approach to different domains of application. Empirical evaluations are conducted to evaluate the feasibility of the proposed method on different task types.
Furthermore, the input modality as well as the target patterns, intensities, colors, and shapes vary with each experiment in order to reflect the flexibility of this conceptual approach.  
The case studies addressed here are as follows:  
\hspace{0.05cm} 1) \textit{Patient survival prediction in ICU }: In this case, the tabular data from electronic health records (EHR) are fed to a network to produce a visualized survival status using distinct colors in a 23x23 image. This case study is a binary classification application. 
\hspace{0.05cm} 2) \textit{Assessing the Diabetic Retinopathy grade}: In this case, the retina \textit{Fundus} images are analyzed by a network to estimate the disease grade and image quality by visualizing intensity and colors in a 23x23 image. This experiment involves a multi-class classification as well as a multi-task application. 
\hspace{0.05cm} 3) \textit{length-of-stay estimation}: The EHR data is fed to a network to estimate visually the length of stay of a patient in which the amount of colored columns in a 45x45 image represents the length of stay. This is a regression case study. 
\hspace{0.05cm} 4) \textit{Diagnosis and prognosis of Alzheimer's Disease (AD) }: In this experiment, multimodal tabular data are fed into a network for producing color-coded AD prediction labels in a colored 23x23 image. This experiment is a multi-class longitudinal prediction.

%%%%%%%%%%%%%%%%%%%%%%%%%%%%%%%%%%%%

\subsection{Experiment 1 - Patient Survival Prediction in ICU} 
In this experiment, colors are used to discriminate among classes. Using data from the first 24 hours of intensive care, this experiment seeks to predict patient survival. The data used in this example is from \textit{WiDS Datathon 2020} challenge \cite{WiDS, WiDS2}. 
This dataset contains tabulated data of electronic health records (EHR) of patients. Survival and death are expressed by 0 or 1, and all the other variables, except for identification information are used as input features, creating a feature vector of 183 values.  
An image with size 23x23x3 (RGB format) is designed as a target image in which a white square symbolizes a label of 0 (survival) and an Orange square symbolizes a label of 1 (warning/fatal). Also, the surrounding pixels are reserved as \textit{RU} areas with zero intensities (black color). 

Figure \ref{fig:survival}B shows the results for a few patients of the testing set. The reported cases from (a) to (f) have estimations close to targets. Case (c) is rather interesting, this means that although the patient survived his symptoms and the information fed as input to the machine algorithm indicates that the subject had a small probability of dying but survived. For case (e), the patient had a high probability of dying and did not survive. Cases (g) and (h) are two cases in which the machine's estimation is not correct. Case (g) is estimated to have a high probability of dying but survived and it is rather a false alarm, and the last case (h) had a small probability of dying and unfortunately died.

%%%%%%%%%%%%%%%%%%%%%%%%%%%%%%%%%%%%
\subsection{Interim Discussion}
Figure \ref{fig:survival}A illustrates the core concept of our proposed VL4ML approach. Our primary goal is to generate medical estimations by utilizing designed target images, and the estimator model is structured to produce these images as its output. Importantly, the medical estimations of interest are inherently encoded within these target images. It is crucial to highlight that, as illustrated in Figure \ref{fig:survival}A, our approach directly integrates the visualization process. Unlike a two-step procedure where predictions are made separately and then visualized, our approach seamlessly combines visualization with the estimation production process. 

Figure \ref{fig:survival}B illustrates the potential of our proposed method. Utilizing the deep ensemble technique, we obtain an average of predictions and the corresponding standard deviation (STD). It is noteworthy that certain predictions result in zero STD images, while brighter STD images signify higher uncertainty. An intriguing observation is that increased uncertainty, as reflected in the STD, is visually apparent in the average of predicted images. Consequently, we will present only the average of predictions in the subsequent experiments.

However, cases (g) and (h) are particularly significant, shedding light on the challenges of uncertainty quantification, especially when it is case-specific rather than system-wide. These cases are instances where human observers perceive them as misclassified, yet the STD of predictions does not significantly differ from cases (d) or (c) which were correctly classified. Hence, obtaining prediction variance through the deep ensemble technique does not provide an immediate solution to the problem. Instead, it can serve as a valuable starting point for establishing a cut-off point and deferring cases for further in-depth analysis.

%%%%%%%%%%%%%%%%%%%%%%%%%%%%%%%%%%%%
\subsection{Experiment 2- Retina Fundus Image Analysis}

This experiment entails a multi-class classification challenge aimed at assessing the severity levels of Diabetic Retinopathy (DR) disease through the analysis of fundus photos. Additionally, the experiment extends to incorporate a multitask scenario, wherein the quality of the input fundus photo is also evaluated.

\begin{figure*}[ht]
\centering
\minipage{\textwidth}
  \includegraphics[width=\linewidth,trim={6cm 4cm 7.5cm 3cm},clip]{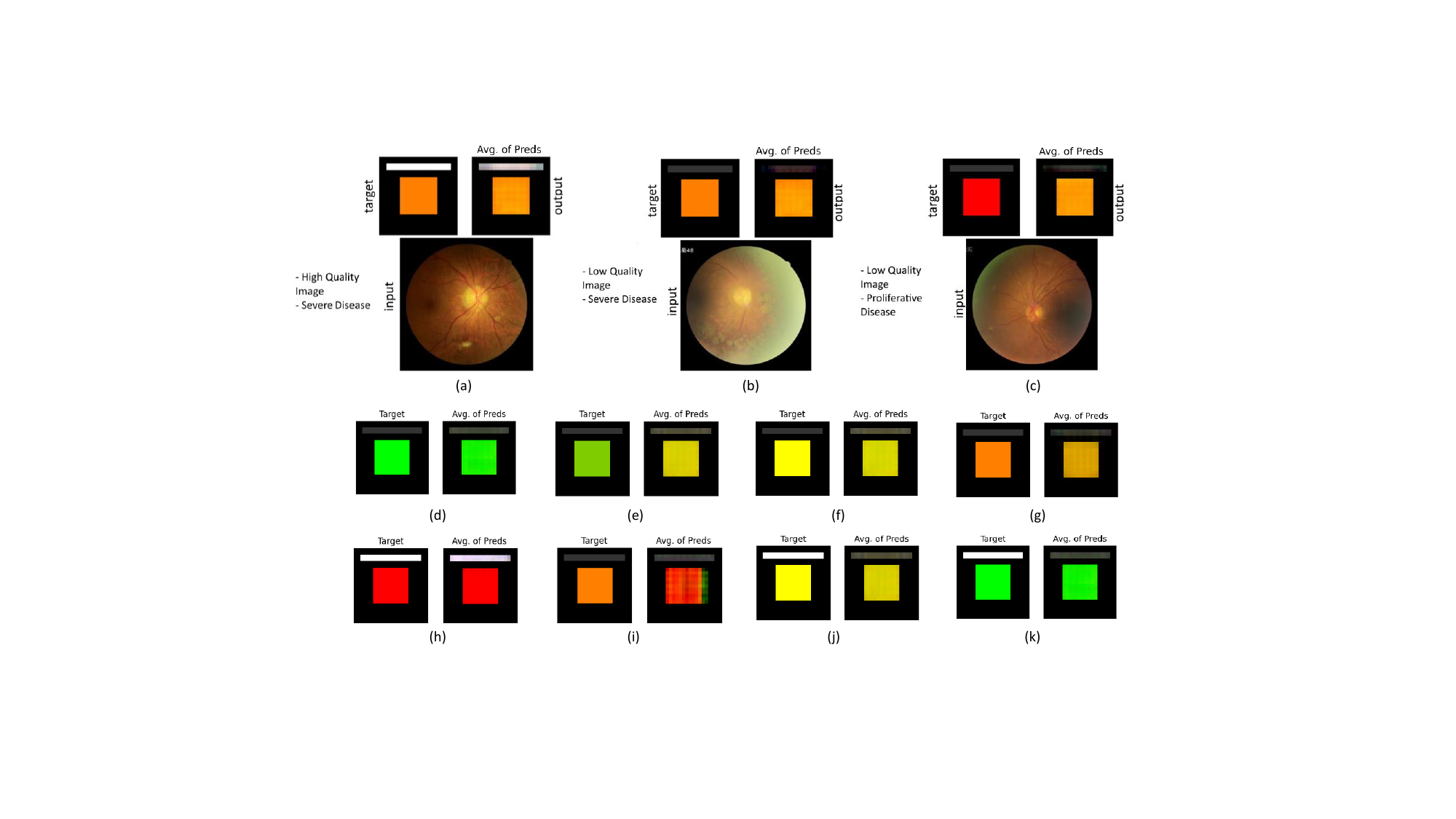}
\endminipage
\caption{Some of the results of the test cases in experiment 2. The left and right images in each sub-figure are the target and achieved output from experiment 2. This is a multi-task scenario including assessing the diabetic retinopathy in 5 levels (No DR, Mild DR, Moderate DR, Severe DR, and proliferative DR) as 5 colors in the center of the image. The quality of input fundus images is also assessed by an upper rectangle (High-quality: White, Low-quality: Dark gray). \\ Short Description: Cases (d) to (h) demonstrate near-accurate estimations, spanning severity levels 0 (No DR) to 4 (Proliferative DR). Notably, all input fundus photos in these cases exhibited poor quality, except for case (h). In contrast, case (i) presents an uncertain output, while cases (j) and (k) showcase accurate severity assessments despite the correct quality assessments.
\footnotesize{ }
}
\label{fig:results_retina}
\end{figure*}

For this experiment, 800 retina Fundus images of 400 subjects from the \textit{DeepDRiD} challenge dataset are used \cite{Deepdrid}. The clinician has rated each image for the severity of diabetic retinopathy on a scale of 0 to 4: 0-No DR, 1-Mild, 2 - Moderate, 3 - Severe, 4 - Proliferative DR. 
Five different intensity levels are color-coded by a square. Indeed as shown in Figure  \ref{fig:results_retina}, the color of the inner square is bright green (RGB(0,1,0)), dark green (RGB(0.5,1,0)), yellow (RGB(1,1,0)), orange(RGB(1,0.5,0)) and red(RGB(1,0,0)) for 0 to 4 levels of disease. To add the quality assessment task, the design includes an additional rectangle whose intensity represents the quality of the Fundus image. As shown in figure \ref{fig:results_retina}, bright white and dark gray rectangles show high-quality and low-quality images, respectively.

%%%%%%%%%%%%%%%%%%%%%%%%%%%%%%%%%%%%

%%%%%%%%%%%%%%%%%%%%%%%%%%%%%%%%%%%%
\subsection{Experiment 3- Estimating the Length-of-Stay in Hospital}

\begin{figure*}[ht]
\centering
\minipage{\textwidth}
  \includegraphics[width=\linewidth,trim={7cm 6.7cm 7cm 6cm},clip]{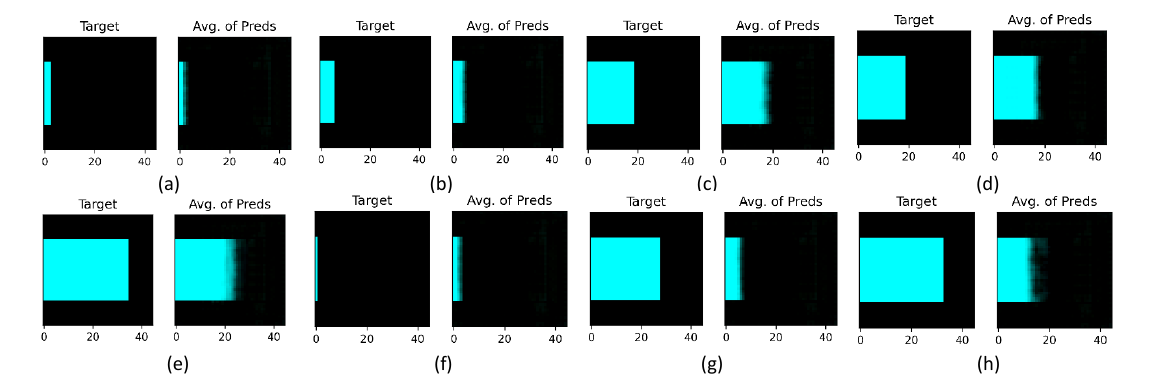}
\endminipage
\caption{Results of several test cases from experiment 3. This is predicting the length-of-stay via coloring the first $M^{th}$ column (started from left) for LOS value equal to $M$. Cases (a) through (s) show the same length of stay in comparison to their ground truth which is on their left side, while cases (e) to (h) seem to have an error in their estimations.
}
\label{fig:results_los}
\end{figure*}

Length-of-stay (LOS) is measured in days between the time of hospital admission and the time of discharge. 
This experiment is a regression problem. The goal of this experiment is to predict the length of stay for each patient at the time of admission visually and the LOS value would be any number between 0 to 40 days. The \textit{MIT MIMIC-III} dataset  \cite{mimic} is used for this experiment due to its availability to the public in the \textit{Physionet} platform \cite{physionet,mimic_physionet}. There are 52 features selected as inputs, and a 45x45x3 target image is constructed, in which each column corresponds to a day as an indicator of time. (i.e. 45 days). Indeed, if the LOS value is $M$, all columns from left to $M^{th}$ column would be in cyan color (RGB(0,1,1)). The 10 rows on top and bottom are colored black to reflect the \textit{RU} area.  
Figure \ref{fig:results_los} shows the results for several cases of the test set.

%%%%%%%%%%%%%%%%%%%%%%%%%%%%%%%%%%%%

%%%%%%%%%%%%%%%%%%%%%%%%%%%%%%%%%%%%

\begin{figure*}[ht]
\centering
\minipage{\textwidth}
  \includegraphics[width=\linewidth,trim={2.5cm 6cm 5.5cm 3.25cm},clip]{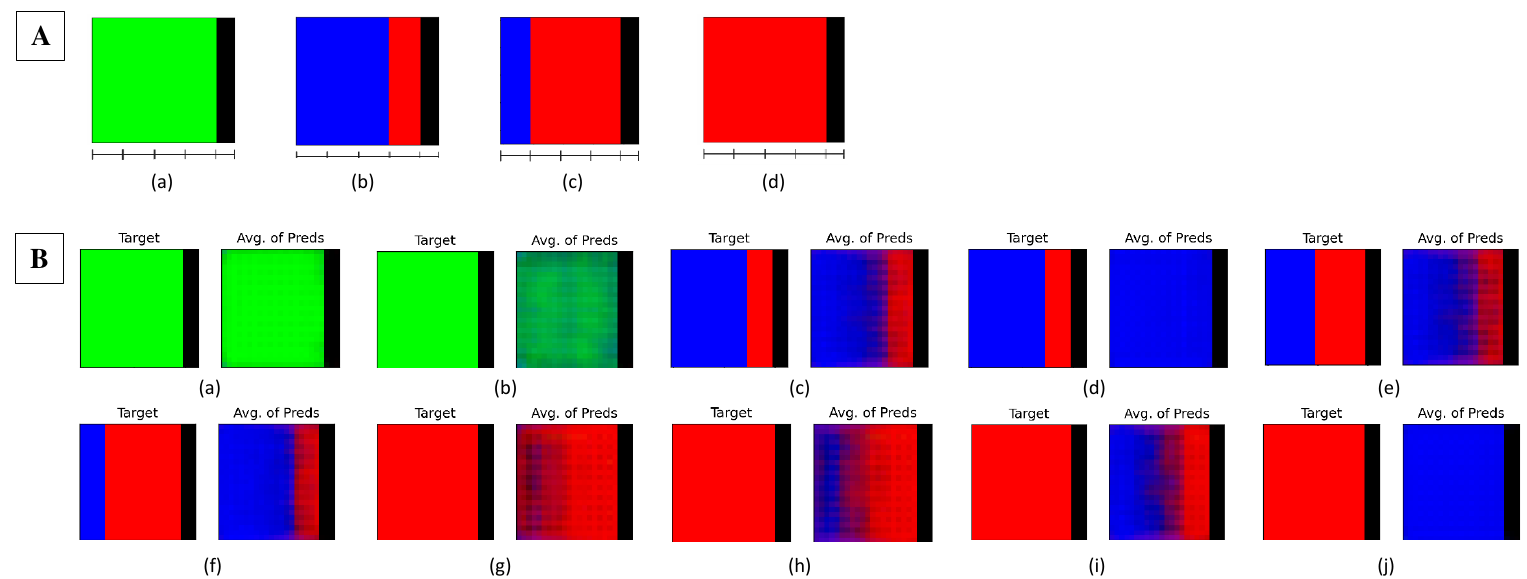}
\endminipage
\caption{A) Examples of the target images for four different disease trajectories. Ruler shows the dedicated areas to time-slots T0 (baseline), T1 ($6^{th}$ month), T2 ($12^{th}$ month), T3 ($24^{th}$ month) and \textit{RU} area, respectively. Notice that, the ruler is not included in the target image. {\footnotesize a) CN from T0 to T3. b) MCI progressed to AD at the T3.  c) MCI progressed to AD at the T1. d) AD from T0 to T3. }
B) Some of the achieved results of the test set in the reported case study for AD progression. The left and right images in each sub-figure are target and predicted output. 
}
\label{fig:results_AD}
\end{figure*}

%%%%%%%%%%%%%%%%%%%%%%%%%%%%%%%%%%%%
\subsection{Experiment 4- Alzheimer's Disease Progression}

In this experiment, we investigate a multiclass classification and prediction of disease progression in Alzheimer's disease in a longitudinal study. The DL model is to produce a visual display with different colored stripes to express the different states of the disease.  The colors chosen in this case are as follows:  \textit{Alzheimer's Disease (AD)}: red, \textit{Mild Cognitive impairment (MCI)}: blue, \textit{Cognitively Normal (CN)}: green.
Therefore, as shown in \ref{fig:results_AD}A, trajectories of cognitive statues over 24 months (including baseline T0 and three referral sessions (T1 ($6^{th}$ month), T2 ($12^{th}$ month), T3 ($24^{th}$ month) ) is demonstrated by a sequence of 4 stripes in which each strip denotes the disease state by its corresponding color. Stripes are followed by a black strip to denote the \textit{RU} area. Figure \ref{fig:results_AD}a shows four examples of desired target images for four different subjects with four different longitudinal trajectories.  

The clinical data used in this experiment is obtained from the \textit{Alzheimer's Disease Neuroimaging Initiative} (\textit{ADNI}) database (\href{http://adni.loni.usc.edu/}{http://ida.loni.usc.edu})
and a total number of 1043 subjects met the trajectories' conditions. 
Multimodal tabular data containing features extracted from \textit{MRI} and \textit{PET} sequences, as well as \textit{demographic information} and \textit{cognitive measurements} were used as input features similar to \cite{neuroimage_sola, frontiers_sola}. 

Figure \ref{fig:results_AD}B demonstrates some of the achieved results of the subjects in test set.
Furthermore, the method seems to provide a more realistic outcome for patients, in particular for longitudinal analysis. For example, while all of the cases (g),(h),(i), and (j) are for patients with steady AD status, the estimated results are different and interesting. Case (g) is diagnosed and labeled correctly, while cases (h) and (i) show different disease progression and Case (j) is a misdiagnosed case with moderate certainty.

\begin{figure}[ht]
    \centering
    \includegraphics[width=0.8\linewidth]{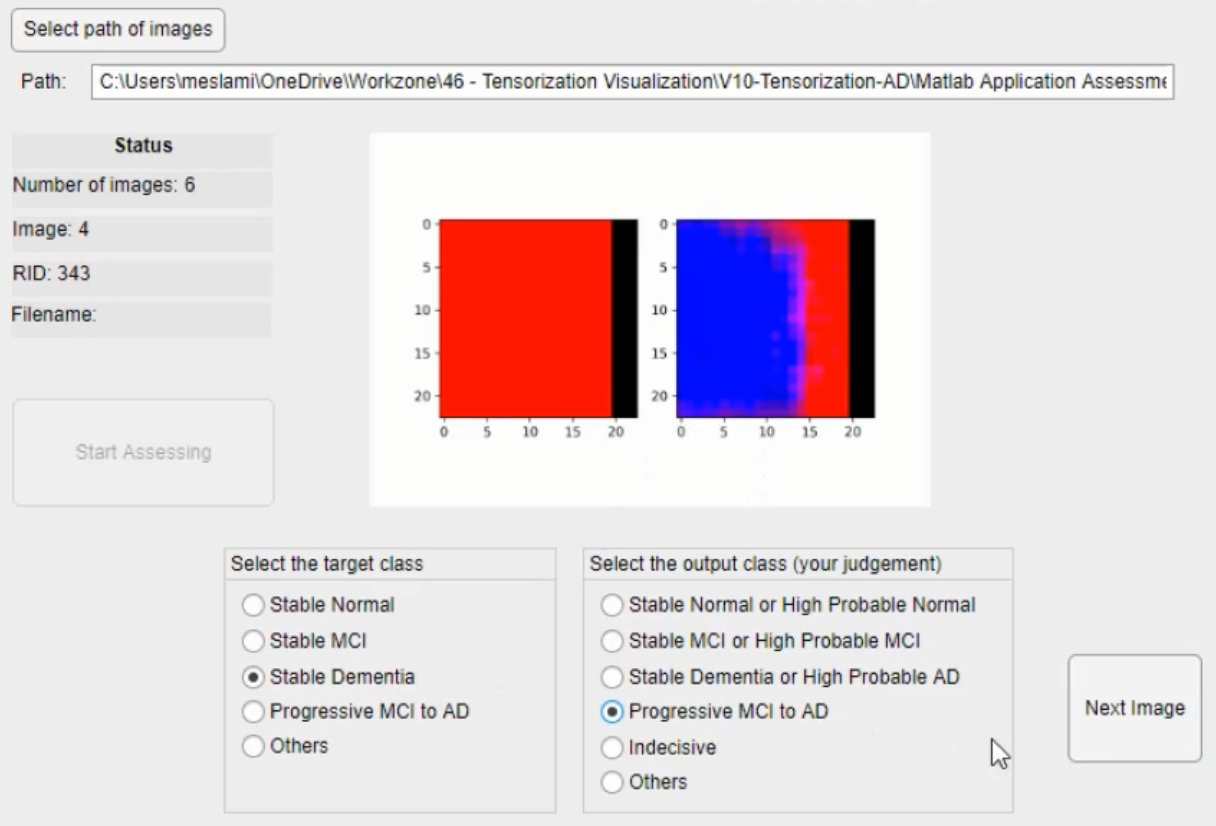}
    \caption{The UI used to assess the experiment 4 results and collect human evaluation.}
    \label{fig:placeholder}
\end{figure}

\subsection{Human Interpretability Evaluation of Visualized Outputs}
\label{Evaluation}

To quantitatively evaluate the proposed framework, we selected the Experiment 4 case study and assessed system performance based on human expert interpretation. Three independent raters evaluated the interpretability of the ML-generated visual outputs. Using a custom MATLAB-based interface (demo: \href{https://youtu.be/yQWFo33RYiQ}{youtube}
), each rater was asked to assign a clinical label based solely on the visual output.
Two classification settings were considered: a 3-class scenario (CN, impaired, others) and a 5-class scenario (CN, MCI, MCIc, AD, others). The “others” category included cases with transitions such as MCI to CN or AD to MCI \cite{eslami2023unique}.

Results (Table \ref{tab:classification_outcomes}) show that, in the 3-class setting, raters achieved an average accuracy of 82\% ± 3\%, indicating that the visual outputs are highly interpretable and preserve clinically meaningful information. In the more granular 5-class setting, accuracy decreased to 68\% ± 5\%, suggesting that interpretability becomes more challenging as task complexity increases. The achieved accuracy is consistent with state-of-the-art literature \cite{eslami2023unique}. These results indicate that clinically meaningful information is preserved in the visual outputs and can be reliably decoded by human observers.

These findings demonstrate that the proposed visual representation enables human observers to recover clinically relevant labels from model outputs, supporting its role as a human-centered explainability mechanism. This evaluation also reflects a practical scenario in which clinicians interpret AI outputs without access to raw probabilities or model internals.

\begin{table}[ht]
    \centering
    \footnotesize % Reducing font size slightly is standard for two-column papers
    \caption{Classification outcomes of experiment 4 as assessed by three raters.}
    \label{tab:classification_outcomes}
    % Use p{3cm} to fix the first column width and force text wrapping
    \begin{tabularx}{\columnwidth}{p{2.5cm} p{1.45cm} p{1.6cm} p{1.6cm}}
        \toprule
        \textbf{Classification Type} & \textbf{Correctly Classified} & \textbf{Misclassified Outcomes} & \textbf{Inconclusive Outcomes} \\ 
        \midrule
        3-Way (CN, impaired, others) & $0.82 \pm 0.03$ & $0.15 \pm 0.004$ & $0.023 \pm 0.002$ \\ 
        \addlinespace
        5-Way (CN, MCI, MCIc, AD, others) & $0.68 \pm 0.05$ & $0.29 \pm 0.01$ & $0.023 \pm 0.002$ \\ 
        \bottomrule
    \end{tabularx}
\end{table}

%% file: sec_survey.tex
\begin{figure*}[ht]
\centering
\minipage{\textwidth}
  \includegraphics[width=\linewidth,trim={1cm 0.8cm 2cm 1cm},clip]{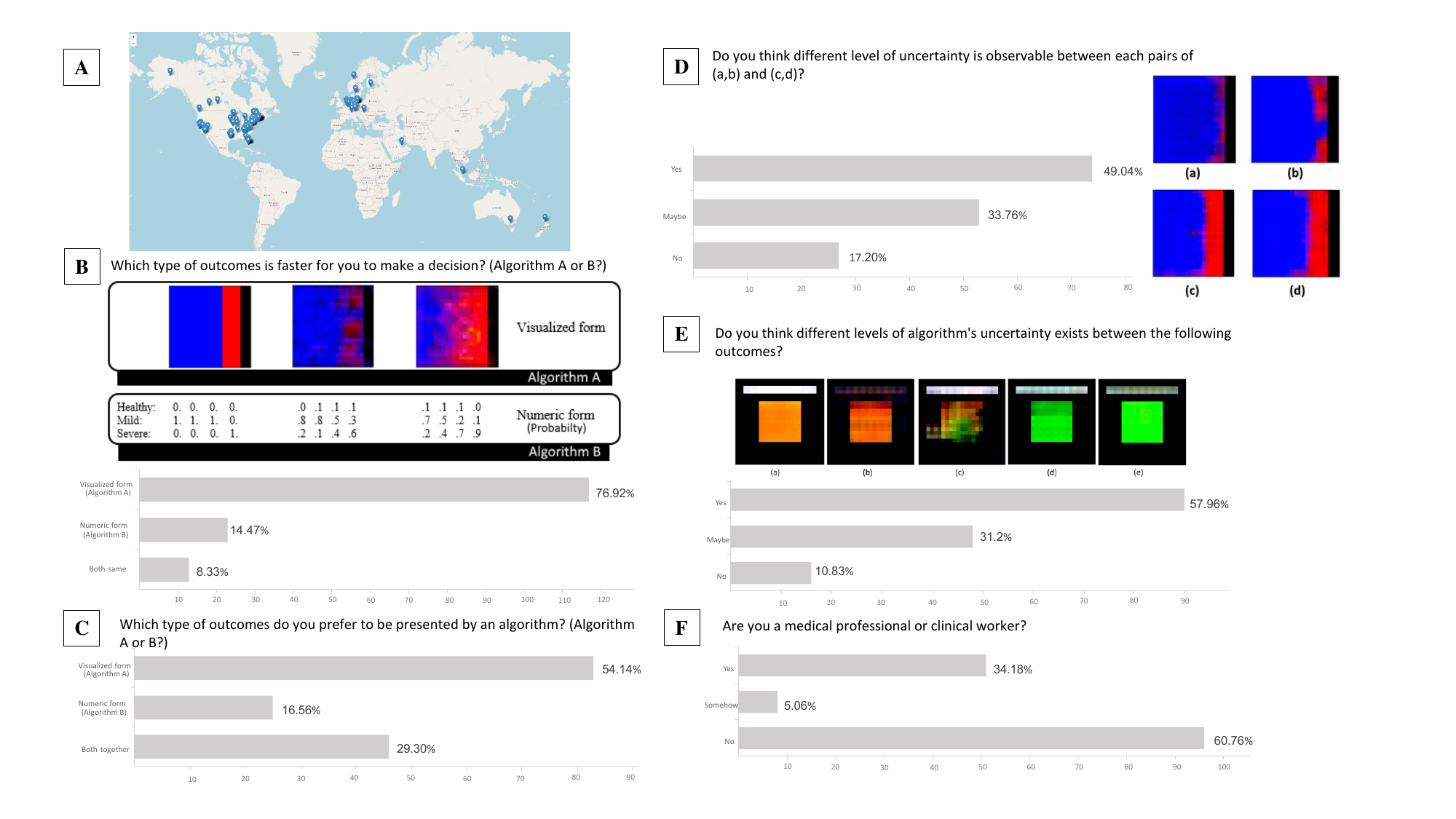}
\endminipage
\caption{ Addressed survey analysis. A) the distribution of people who participated. B-F) Five exemplary questions and achieved results. The full survey and results are available in the appendix. A total of 155 individuals participated in this survey study, and the horizontal axis in the barplots represents counts.
}
\label{fig:survey}
\end{figure*}

An online survey has been created with the Qualtrics platform for collecting users' feedback
and the objective was to assess user preference in the ability to interpret and understand the visual display. 
%It is also an assessment of how easy it is to remember the visual display and how easy in making a decision, especially when the outcome is ambiguous and may be infused with some degree of uncertainty. 
The survey assessed participants' preferences, ease of interpretation, memorability, decision speed, and uncertainty
perception when comparing visual output representations (Algorithm~A) against conventional numeric probability outputs (Algorithm~B) across three clinical tasks: ICU survival prediction, Alzheimer's disease progression, and Multitask diabetic retinopathy severity assessment.
A few initial samples were given to participants to familiarize them with the meaningfulness of the visualization format and its color coding. %Ten questions related to three application of survival analysis, disease progression estimation and multitask disease severity estimation were posed in this survey.

This study was determined to be exempt from full Institutional Review Board (IRB) review by the Mass General Brigham Institutional Review Board in accordance with 45 CFR 46.104(d)(2). The study involved anonymous online survey responses collected from adult volunteers, with no collection of personally identifiable information, no clinical intervention, no patient recruitment, and no biological sample collection. All participants were informed of the study purpose and provided consent before completing the survey. 
Survey responses were collected between February 1, 2024, and March 21, 2024.

We shared the survey on Facebook Ads, LinkedIn, and among various groups within messaging apps. 158 individuals responded to the survey and Figure \ref{fig:survey}a shows the distribution of collected votes. Five exemplary questions and their corresponding results are also shown in Fig. \ref{fig:survey}. 
The full survey and results are available in the appendix. 

The analysis goes beyond the descriptive statistics and introduces (i)~inferential tests with effect-size
estimates and Bonferroni-corrected $p$-values, (ii)~subgroup comparisons across clinical background, age, and gender,
(iii)~a Spearman correlation analysis across all ordinal responses,
(iv)~a task-complexity analysis, (v)~a preference-consistency test,
and (vi)~a logistic regression identifying predictors of visual
preference. 

\subsection{Participant Demographics}
Understanding the composition of the survey sample is essential for
interpreting all subsequent results and for assessing the
generalisability of the findings. A sample that is heavily skewed
toward a particular age group or professional background may limit how
broadly the conclusions can be applied to real clinical populations.

\begin{figure}[]
  \centering
  \includegraphics[width=\linewidth]{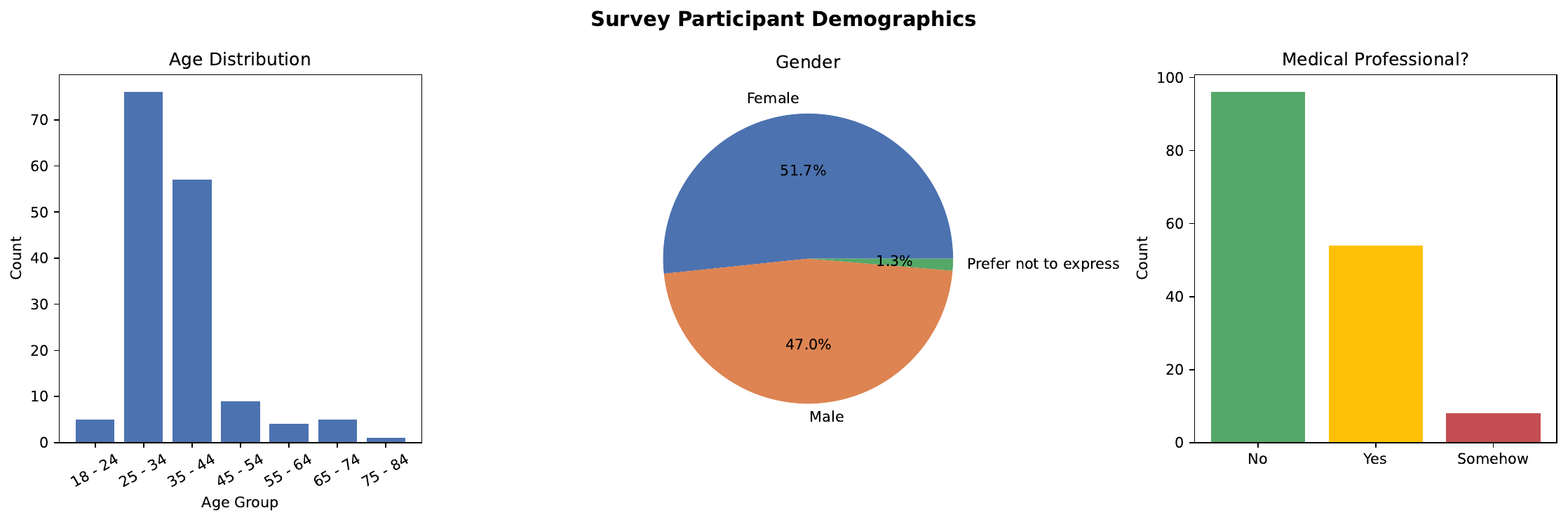}
  \caption{Demographic distribution of the 158 survey participants.
    Left: age group; Centre: gender; Right: clinical background.}
  \label{fig:demographics}
\end{figure}

The survey attracted 158 completed responses distributed across multiple
countries (Figures \ref{fig:survey}A, ~\ref{fig:demographics}). The sample was approximately gender-balanced: 51.7\% female ($n=78$) and 47.0\% male ($n=71$). The majority of participants (84.7\%, $n=133$) were between 25 and 44 years of age, reflecting the primary demographic reached through the social-media and professional-network recruitment strategy.
Regarding clinical background, 34.2\% ($n=54$) identified as medical professionals or clinical workers, 5.1\% ($n=8$) selected ``Somewhat'', and 60.8\% ($n=96$) had no clinical background. Among those who identified as clinical workers, 82.5\% ($n=33$ out of 40 who answered the follow-up question) were physicians.

The sample is sufficiently diverse to support the analyses that follow,
with a meaningful proportion (39.2\%) of clinical or medical
participants providing domain-relevant perspectives alongside the
general public. The age skew toward younger adults is consistent with
online recruitment and should be kept in mind when interpreting the
logistic regression results (Section~\ref{sec:logreg}).

\subsection{Overall Response Distributions}
\begin{figure}[]
  \centering
  \includegraphics[width=\linewidth]{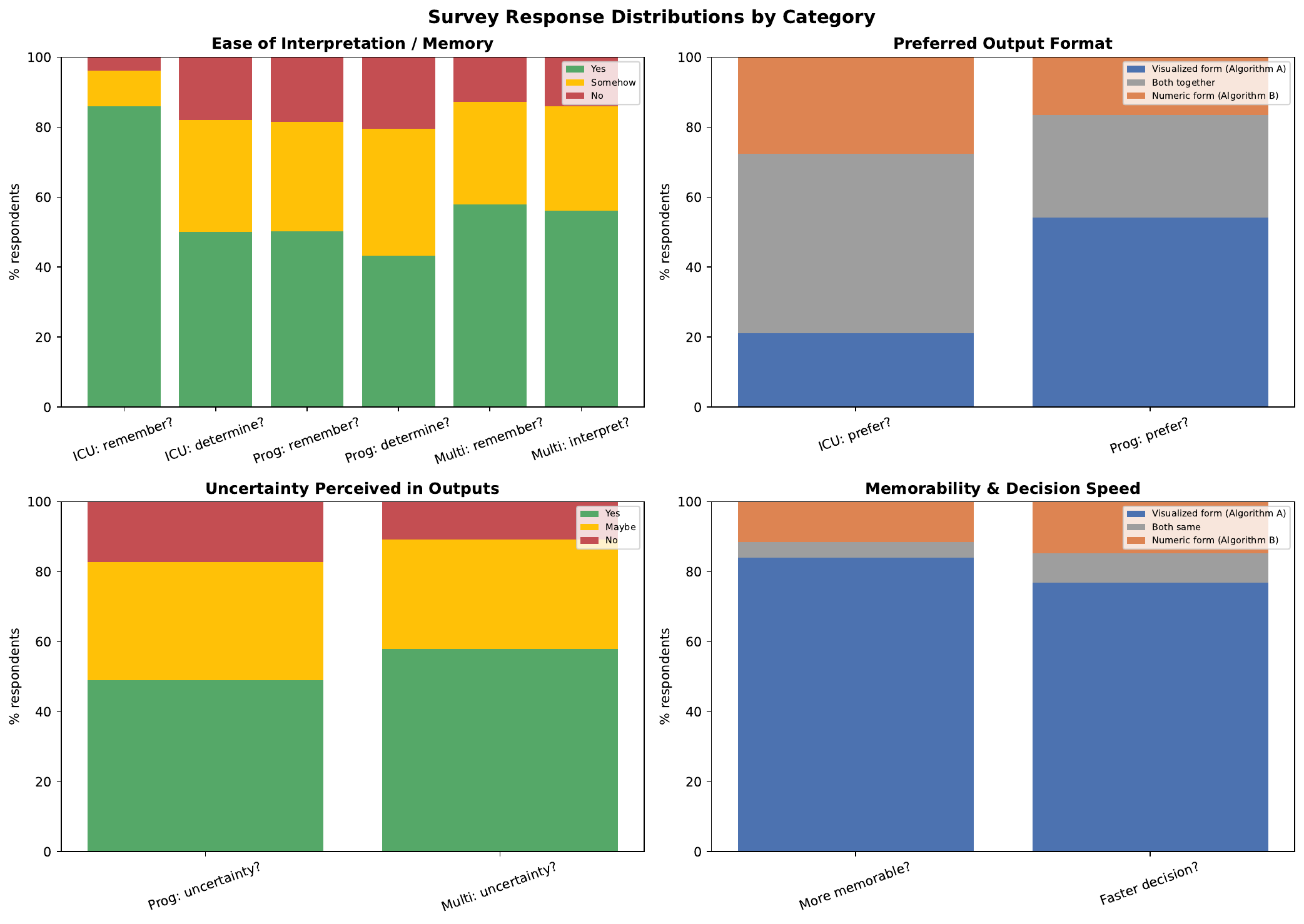}
  \caption{Stacked bar distributions across four response categories.
    Top-left: ease of interpretation and memory per task;
    Top-right: preferred output format per task;
    Bottom-left: uncertainty perception per task;
    Bottom-right: memorability and decision speed.}
  \label{fig:distributions}
\end{figure}

Before conducting any inferential tests, it is important to characterize
the raw distribution of responses across all question categories.
Stacked bar charts provide an at-a-glance view of where the majority
of opinion lies for each dimension (ease, preference, memorability, and
uncertainty), which informs which comparisons are likely to be
statistically meaningful and which are floor- or ceiling-affected. 
Figure~\ref{fig:distributions} summarizes the four main question
categories as stacked bar charts.

The visual format is strongly favoured for memorability (84.0\%) and
decision speed (76.9\%). Preference for visual output as the
\emph{sole} format is more modest for the simpler ICU binary task
(21.2\%), with most participants preferring both formats together
(51.3\%), suggesting that visual outputs complement rather than replace
numeric ones for simple binary decisions. For the more complex
disease-progression task, however, 54.1\% preferred the visual format
alone. Uncertainty was perceived by the majority in both tasks
($\geq$82.8\% answered Yes or Maybe), supporting the utility of the
VL4ML uncertainty communication mechanism.

To characterise the overall reception of the VL4ML visual format, we
adopt a binary ``positive response'' criterion that merges
\textit{Yes} and \textit{Somehow} (or \textit{Maybe} for uncertainty
items) into a single affirmative category, following standard
practice in Likert-scale reporting. Under this
criterion, Table~\ref{tab:binary_descriptive} shows that positive
reception exceeded \textbf{79\%} across every question and every
clinical task, reaching \textbf{96.2\%} for memorability of the
ICU survival display.  
Ease of interpretation was highest for the binary ICU task
(82.1\%--96.2\%), reflecting the simplicity of a two-colour
white/orange encoding. Despite the increased complexity of the
disease-progression and multitask displays, positive ease ratings
remained above 79\% in all cases, demonstrating that VL4ML outputs
are accessible even for multi-class and longitudinal predictions.
Uncertainty perception was affirmed by 82.8\% and 89.2\% of
respondents for the disease-progression and multitask tasks,
respectively, confirming that ensemble deviation visualization effectively communicates
model confidence to a general audience.

\begin{table*}[ht]
\centering
\scriptsize
\caption{Binary descriptive statistics for ease-of-interpretation and uncertainty-perception questions. 
  %Positive response = \textit{Yes}  $+$ \textit{Somehow} (ease) or \textit{Yes} $+$ \textit{Maybe} (uncertainty). $N \approx 157$ per item.
  }
\label{tab:binary_descriptive}
\renewcommand{\arraystretch}{1.3}
\setlength{\tabcolsep}{6pt}
\begin{tabular}{llcccc}
\toprule
\textbf{Task} & \textbf{Question} &
  \textbf{Yes (\%)} & \textbf{Partial (\%)} & \textbf{No (\%)} &
  \textbf{Positive (\%)}  \\
\midrule
 
\multirow{2}{*}{ICU Survival}
  & Easy to remember?       & 86.0 & 10.2 &  3.8 & \textbf{96.2}  \\
  & Easy to determine?      & 50.0 & 32.1 & 17.9 & \textbf{82.1}  \\
\midrule
 
\multirow{3}{*}{\shortstack[l]{Disease\\Progression}}
  & Easy to remember?       & 50.3 & 31.2 & 18.5 & \textbf{81.5}  \\
  & Easy to determine?      & 43.3 & 36.3 & 20.4 & \textbf{79.6}  \\
  & Uncertainty observable? & 49.0 & 33.8 & 17.2 & \textbf{82.8}  \\
\midrule
 
\multirow{3}{*}{\shortstack[l]{Multitask\\Severity}}
  & Easy to remember?       & 58.0 & 29.3 & 12.7 & \textbf{87.3}  \\
  & Easy to interpret?      & 56.1 & 29.9 & 14.0 & \textbf{86.0}  \\
  & Uncertainty observable? & 58.0 & 31.2 & 10.8 & \textbf{89.2}  \\
 
\bottomrule
\multicolumn{6}{l}{\footnotesize Partial = \textit{Somehow} (ease) or \textit{Maybe} (uncertainty).
  Positive = Yes $+$ Partial.} \\
\end{tabular}
\end{table*}

\subsection{Clinician vs.\ Non-Clinician Subgroup Comparison}

The most critical subgroup comparison for a clinical AI paper is whether
medical professionals respond differently to visual outputs than
laypeople. If clinicians — who are accustomed to numerical readouts and
probability-based reasoning — show lower preference for or
interpretability of visual outputs, it would signal a barrier to
real-world adoption. Conversely, if responses are consistent across
groups, it strengthens the claim that VL4ML is universally accessible.

\begin{figure}[]
  \centering
  \includegraphics[width=\linewidth]{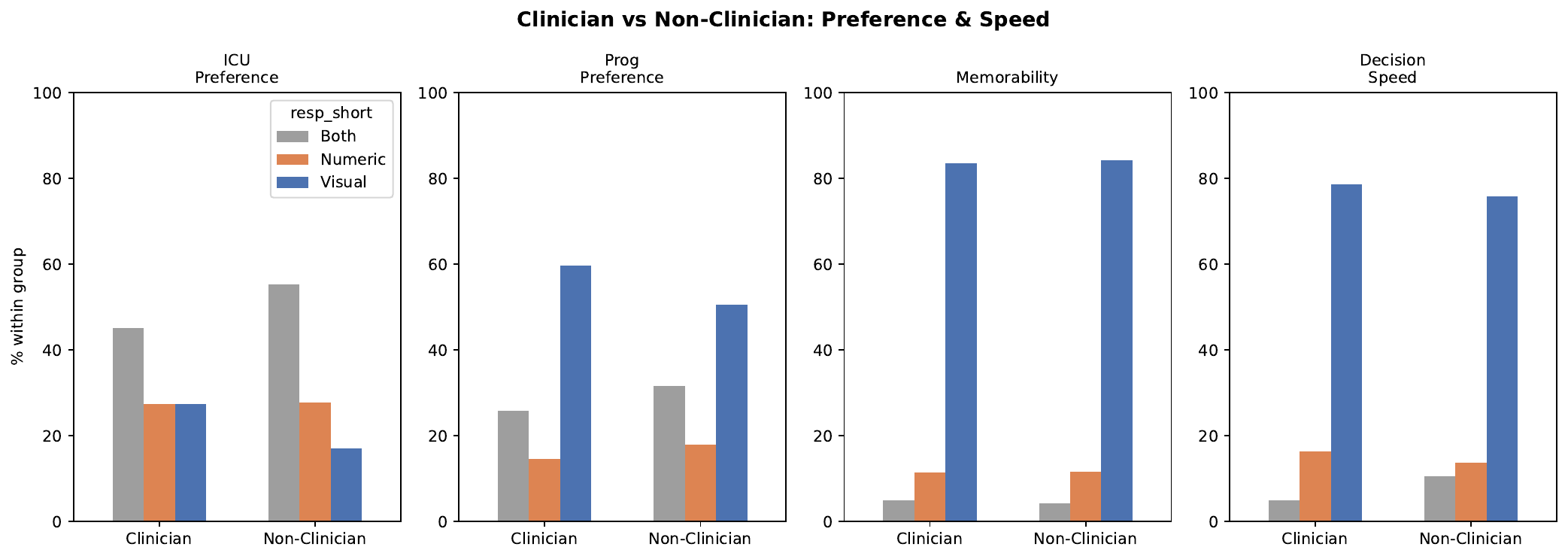}
  \caption{Preference and speed distributions stratified by clinical
    background across four question dimensions. Percentages are
    computed within each group.}
  \label{fig:clin_pref}
\end{figure}

\begin{figure}[]
  \centering
  \includegraphics[width=\linewidth]{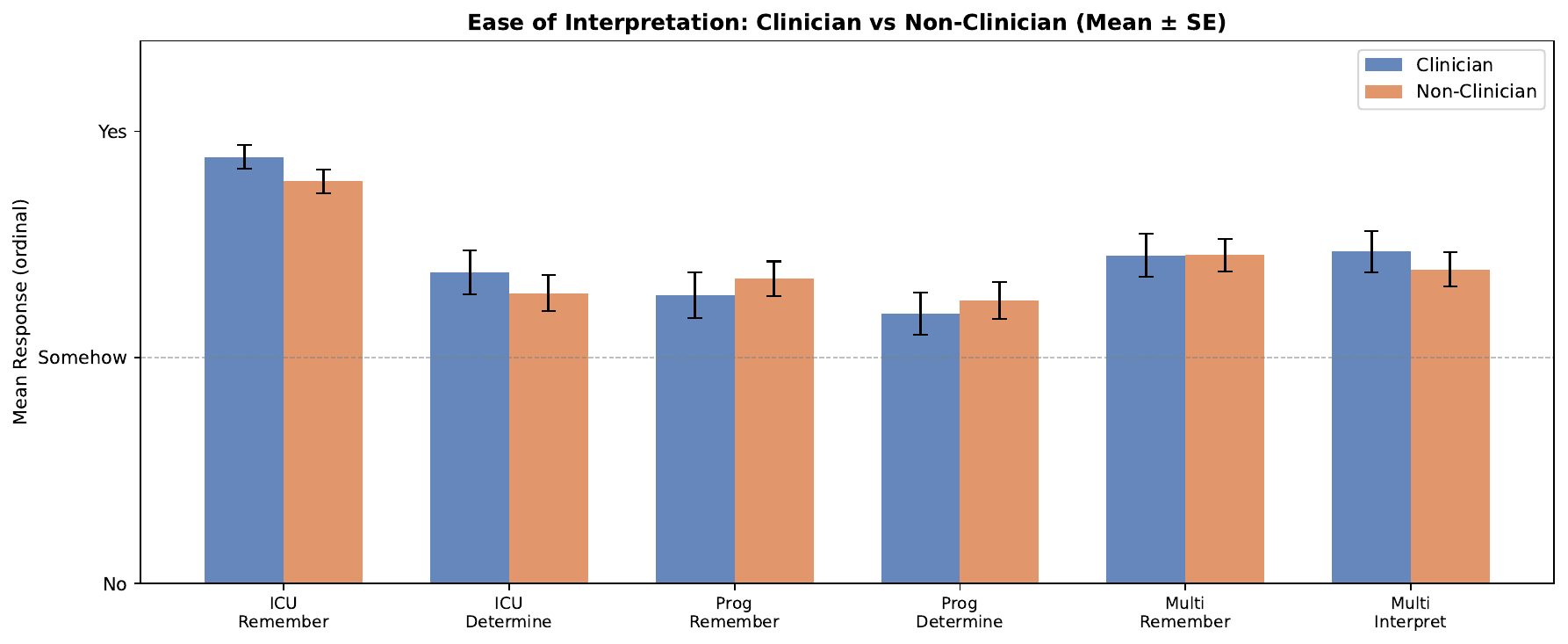}
  \caption{Mean ease-of-interpretation scores ($\pm$SE) for clinicians
    and non-clinicians across all six ease questions. Ordinal scale:
    1 = No, 2 = Somehow, 3 = Yes. Dashed line indicates the
    ``Somehow'' midpoint. No pairwise difference reached significance
    after Bonferroni correction.}
  \label{fig:clin_ease}
\end{figure}

Chi-square tests ($\chi^2$) were conducted for all 12 survey questions,
comparing clinicians ($n=62$) against non-clinicians ($n=96$). Cramer's~$V$
was computed as the effect-size measure. Bonferroni correction was
applied ($\alpha_{\text{corrected}} = 0.05 / 12 = 0.004$).
Results are shown in Table~\ref{tab:clinician}.
Figure~\ref{fig:clin_pref} displays the grouped preference
distributions, and Figure~\ref{fig:clin_ease} shows mean ease scores
with standard errors, stratified by clinical background.

\begin{table}[]
\centering
\caption{Chi-square test results: Clinician vs.\ Non-Clinician.
  $V$ = Cramer's~$V$ effect size.
  All $p_{\text{Bonf}}$ values exceed 0.05; ``ns'' = not significant.}
\label{tab:clinician}
\renewcommand{\arraystretch}{1.2}
\scriptsize
\begin{tabular}{lccccl}
\toprule
\textbf{Question} & $\chi^2$ & \textit{df} & $p$ & $p_{\text{Bonf}}$ & $V$ \\
\midrule
ICU — preferred format       & 2.66 & 2 & 0.264 & 1.000 & 0.065 \\
Prog — preferred format      & 1.27 & 2 & 0.531 & 1.000 & 0.000 \\
Prog — memorability          & 0.04 & 2 & 0.979 & 1.000 & 0.000 \\
Prog — decision speed        & 1.63 & 2 & 0.443 & 1.000 & 0.000 \\
Prog — uncertainty visible   & 0.03 & 2 & 0.987 & 1.000 & 0.000 \\
Multi — uncertainty visible  & 0.96 & 2 & 0.620 & 1.000 & 0.000 \\
ICU — easy to remember       & 3.40 & 2 & 0.183 & 1.000 & 0.094 \\
ICU — easy to determine      & 0.67 & 2 & 0.714 & 1.000 & 0.000 \\
Prog — easy to remember      & 0.43 & 2 & 0.806 & 1.000 & 0.000 \\
Prog — easy to determine     & 1.45 & 2 & 0.485 & 1.000 & 0.000 \\
Multi — easy to remember     & 0.73 & 2 & 0.693 & 1.000 & 0.000 \\
Multi — easy to interpret    & 0.55 & 2 & 0.761 & 1.000 & 0.000 \\
\bottomrule
\end{tabular}
\end{table}

None of the 12 tests reached significance after Bonferroni correction
(all $p_{\text{Bonf}} = 1.000$, all $V < 0.10$). The mean ease scores
between clinicians and non-clinicians were virtually identical across
all tasks (e.g., Multitask ease: $M_{\text{clin}} = 2.45$,
$M_{\text{non-clin}} = 2.45$). This is a \textbf{strong positive
finding}: visual outputs are equally interpretable and equally preferred
regardless of medical expertise, suggesting that VL4ML does not require
domain-specific training to use effectively and is accessible to both
patients and clinical staff.

\subsection{ Age and Gender Subgroup Comparison}

Age is a known moderator of technology acceptance and visual literacy.
Older adults may find novel visual encodings less intuitive, while
younger cohorts raised on data-rich interfaces may adapt more readily.
If strong age effects exist, targeted design adaptations or onboarding
procedures may be warranted for different patient demographics.

Chi-square tests compared three broad age bands: 18--34 ($n=81$),
35--54 ($n=66$), and 55$+$ ($n=10$). Results are summarized in
Table~\ref{tab:age}.

\begin{table}[]
\centering
\caption{Chi-square test results: Age groups (18--34 / 35--54 / 55+).
  One item reached significance after Bonferroni correction (marked~*).}
\label{tab:age}
\renewcommand{\arraystretch}{1.2}
\scriptsize
\begin{tabular}{lccccl}
\toprule
\textbf{Question} & $\chi^2$ & \textit{df} & $p$ & $p_{\text{Bonf}}$ & $V$ \\
\midrule
ICU — preferred format       &  2.68 & 4 & 0.613 & 1.000 & 0.000 \\
Prog — preferred format      & 11.11 & 4 & 0.025 & 0.305 & 0.151 \\
Prog — memorability          & 15.85 & 4 & 0.003 & \textbf{0.038} & 0.196 \\
Prog — decision speed        &  6.22 & 4 & 0.183 & 1.000 & 0.084 \\
Prog — uncertainty visible   &  4.71 & 4 & 0.318 & 1.000 & 0.047 \\
Multi — uncertainty visible  &  2.52 & 4 & 0.642 & 1.000 & 0.000 \\
ICU — easy to remember       &  6.00 & 4 & 0.199 & 1.000 & 0.080 \\
ICU — easy to determine      & 10.01 & 4 & 0.040 & 0.482 & 0.139 \\
Prog — easy to remember      &  3.65 & 4 & 0.456 & 1.000 & 0.000 \\
Prog — easy to determine     &  3.03 & 4 & 0.552 & 1.000 & 0.000 \\
Multi — easy to remember     &  2.92 & 4 & 0.572 & 1.000 & 0.000 \\
Multi — easy to interpret    & 10.63 & 4 & 0.031 & 0.372 & 0.146 \\
\bottomrule
\end{tabular}
\end{table}

Only one question — memorability of the disease-progression visual
($\chi^2(4)=15.85$, $p_{\text{Bonf}}=0.038$, $V=0.196$) — reached
significance after correction. This medium-small effect suggests that
younger participants found the colour-coded longitudinal display slightly
more memorable, which is consistent with literature on picture-superiority
effects being more pronounced in younger adults. Overall, the findings
indicate that age has minimal impact on VL4ML interpretability,
though the small $n$ for the 55$+$ group (only 10 participants) limits
the statistical power of this comparison and warrants a dedicated study
with older adults.

Gender-related differences in visual-spatial processing and health
information preferences have been reported in the literature. Checking
whether the VL4ML outputs are perceived differently across genders
supports an inclusive design claim and may reveal whether the colour
and pattern encoding choices are equitable.
Chi-square tests compared male ($n=71$) and female ($n=78$)
participants. No significant effects were found after Bonferroni
correction (all $p_{\text{Bonf}} > 0.05$, all $V < 0.17$). The
largest (non-significant) effect was for ``ICU — easy to determine''
($\chi^2(2)=5.78$, $p=0.056$, $V=0.160$).
Gender does not significantly moderate the interpretation of or
preference for VL4ML outputs. This finding supports the
generalisability of the framework across a diverse patient population
and is important for claims of equitable AI design.

\subsection{Effect-Size Overview Heatmap}

Presenting effect sizes (Cramer's~$V$) alongside $p$-values in a
unified visualisation allows readers to compare the \emph{magnitude}
of group differences at a glance, which is more informative than
significance flags alone. Small $p$-values with trivial effect sizes
carry a different meaning than moderate $p$-values with meaningful
effect sizes, especially in small-$n$ subgroups.

\begin{figure}[]
  \centering
  \includegraphics[width=\linewidth]{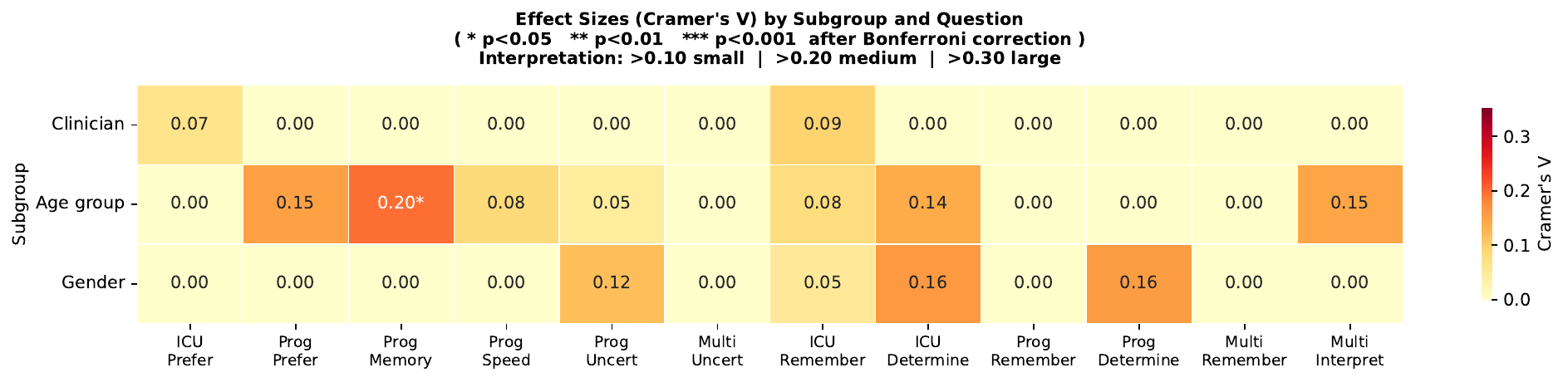}
  \caption{Heatmap of Cramer's~$V$ effect sizes across subgroup
    comparisons (rows) and survey questions (columns). 
    %Cells include the $V$ value and Bonferroni-corrected significance stars (* $p{<}0.05$, ** $p{<}0.01$, *** $p{<}0.001$).
    Effect-size benchmarks: $V{>}0.10$ small, $V{>}0.20$ medium,    $V{>}0.30$ large.
    }
  \label{fig:heatmap}
\end{figure}

Figure~\ref{fig:heatmap} shows Cramer's~$V$ for all three subgroup
comparisons and all 12 questions. Bonferroni-corrected significance
stars are embedded in each cell. The heatmap reveals a consistently low effect-size landscape across all
subgroups, with the single exception of age-related memorability
($V=0.196$). The near-zero Cramer's~$V$ values for the clinician and
gender comparisons confirm that VL4ML is equally effective across these
demographic dimensions. This consolidated view is particularly useful
for communicating null-but-meaningful subgroup findings in a concise,
publication-ready format.

\subsection{Task-Complexity Analysis}

VL4ML is applied across tasks of varying complexity: a simple binary
classification (ICU survival), a multi-class longitudinal prediction
(disease progression), and a concurrent multi-task scenario
(retinopathy severity + image quality). If ease of interpretation
degrades significantly with complexity, it suggests the visual encoding
strategy needs further refinement for complex outputs; a key direction
for future work.

\begin{figure}[]
  \centering
  \includegraphics[width=0.85\linewidth]{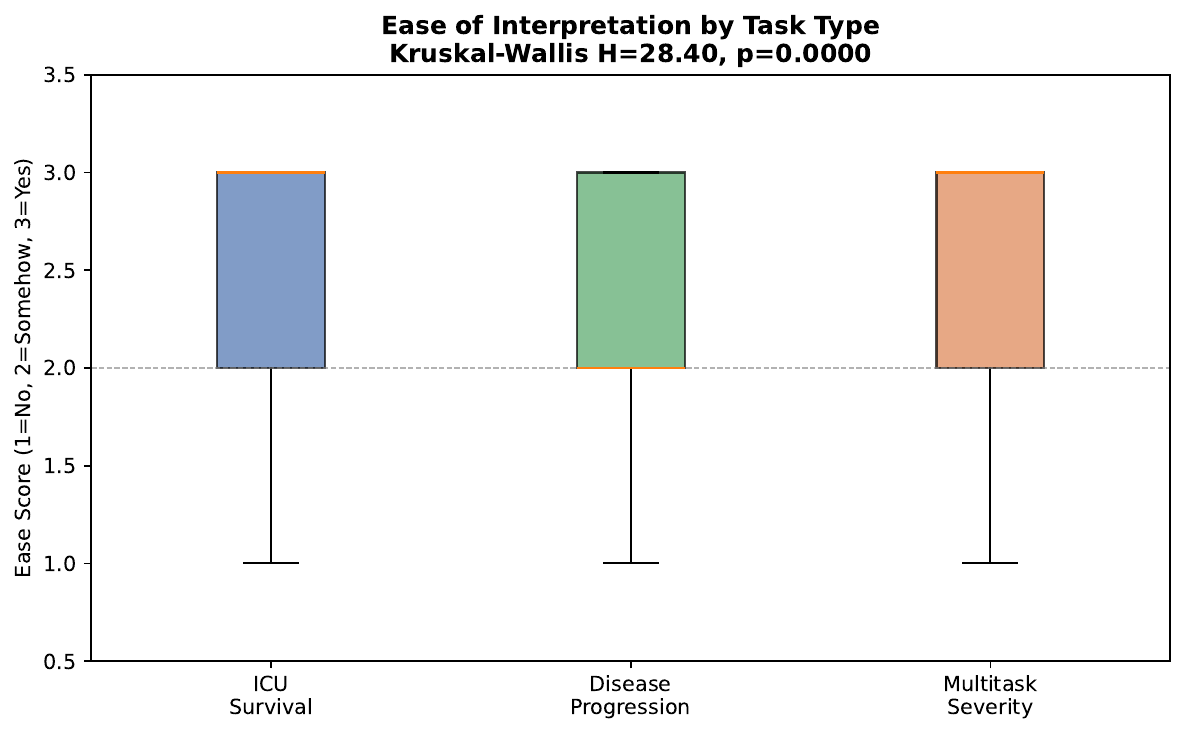}
  \caption{Box plots of ease-of-interpretation scores across three
    task types. Ordinal scale: 1 = No, 2 = Somehow, 3 = Yes.
    Dashed line marks the ``Somehow'' midpoint. Kruskal--Wallis
    $H = 28.40$, $p < 0.001$.}
  \label{fig:tasks}
\end{figure}

A Kruskal--Wallis test compared ease-of-interpretation ordinal scores
(pooling the two ease questions per task) across the three tasks:
ICU ($n=313$), Disease Progression ($n=314$), and Multitask ($n=314$).
The test was significant ($H = 28.40$, $p < 0.001$), and Bonferroni-corrected
post-hoc Mann--Whitney~U tests (threshold $\alpha = 0.017$) revealed
significant pairwise differences between all three tasks
(Table~\ref{tab:tasks}). Figure~\ref{fig:tasks} shows the
corresponding box plots.

\begin{table}[]
\centering
\caption{Post-hoc Mann--Whitney U comparisons of ease scores across
  task types (Bonferroni threshold $\alpha = 0.017$).
  $r$ = rank-biserial correlation (effect size). All comparisons
  significant.}
\label{tab:tasks}
\renewcommand{\arraystretch}{1.2}
\footnotesize
\begin{tabular}{llcccc}
\toprule
\textbf{Task A} & \textbf{Task B} & $U$ & $p$ & $r$ & Sig. \\
\midrule
ICU       & Progression & 59{,}788 & $<$0.001 & $-$0.217 & * \\
ICU       & Multitask   & 54{,}374 &   0.007  & $-$0.106 & * \\
Progression & Multitask &  43{,}664 &   0.006  &   0.114 & * \\
\bottomrule
\end{tabular}
\end{table}

Ease of interpretation decreases as task complexity increases:
the ICU binary task is easiest (highest ease scores), disease
progression is hardest (most complex longitudinal colour stripes),
and the multitask scenario falls between them. The effect sizes are
small-to-moderate ($|r| = 0.11$--$0.22$), indicating that while
statistically significant, the practical degradation in usability is
modest. This is encouraging for VL4ML's scalability, but underlines
the importance of careful visual design for multi-class or longitudinal
outputs. Future work should investigate adaptive or interactive
representations that reduce cognitive load for complex tasks.

\subsection{Preference Consistency Across Tasks}

If a participant who prefers the visual format in the ICU task also
prefers it in the disease-progression task, it suggests that preference
is a stable individual trait rather than being purely task-driven.
This has important design implications: a single preference elicitation
at the start of a clinical session could be used to personalise the
display format for the remainder of the encounter.

Cross-tabulating the preferred output format between the ICU task (Q4)
and the disease-progression task (Q9) for 156 participants who answered
both questions showed that 49.4\% ($n = 77$) chose the \emph{same}
format in both tasks. A chi-square test of the $3 \times 3$
cross-tabulation was highly significant ($\chi^2(4) = 41.22$, $p <
0.001$, Cramer's $V = 0.347$), confirming a strong association between
preferences across the two tasks (Table~\ref{tab:consistency}).

\begin{table}[]
\centering
\caption{Cross-tabulation of preferred output format: ICU task (Q4, rows)
  vs.\ Disease Progression task (Q9, columns). $n = 156$.}
\label{tab:consistency}
\renewcommand{\arraystretch}{1.2}
\scriptsize
\begin{tabular}{lccc}
\toprule
 & \textbf{Visual (Q9)} & \textbf{Both (Q9)} & \textbf{Numeric (Q9)} \\
\midrule
\textbf{Both (Q4)}    & 43 & 34 &  3 \\
\textbf{Numeric (Q4)} & 17 &  7 & 19 \\
\textbf{Visual (Q4)}  & 24 &  5 &  4 \\
\bottomrule
\multicolumn{4}{l}{$\chi^2(4) = 41.22$, $p < 0.001$, Cramer's $V = 0.347$} \\
\end{tabular}
\end{table}

Preference for output format is a moderately stable individual
characteristic ($V = 0.35$, a medium-large effect), not purely
task-dependent. This supports the feasibility of a one-time preference
assessment at patient onboarding to guide personalized AI output
formatting throughout a clinical session. It also explains why
aggregate preference statistics are meaningful: they reflect genuine
inter-individual variation rather than noise.

\subsection{Uncertainty Perception}
A core design goal of VL4ML is to communicate model uncertainty
perceptually through the visual output. If users cannot distinguish
uncertain from confident predictions visually, this design goal fails,
regardless of how well the average-prediction images look. This analysis
checks whether uncertainty is actually perceived and whether clinicians
are better at detecting it.

\begin{figure}[]
  \centering
  \includegraphics[width=0.85\linewidth]{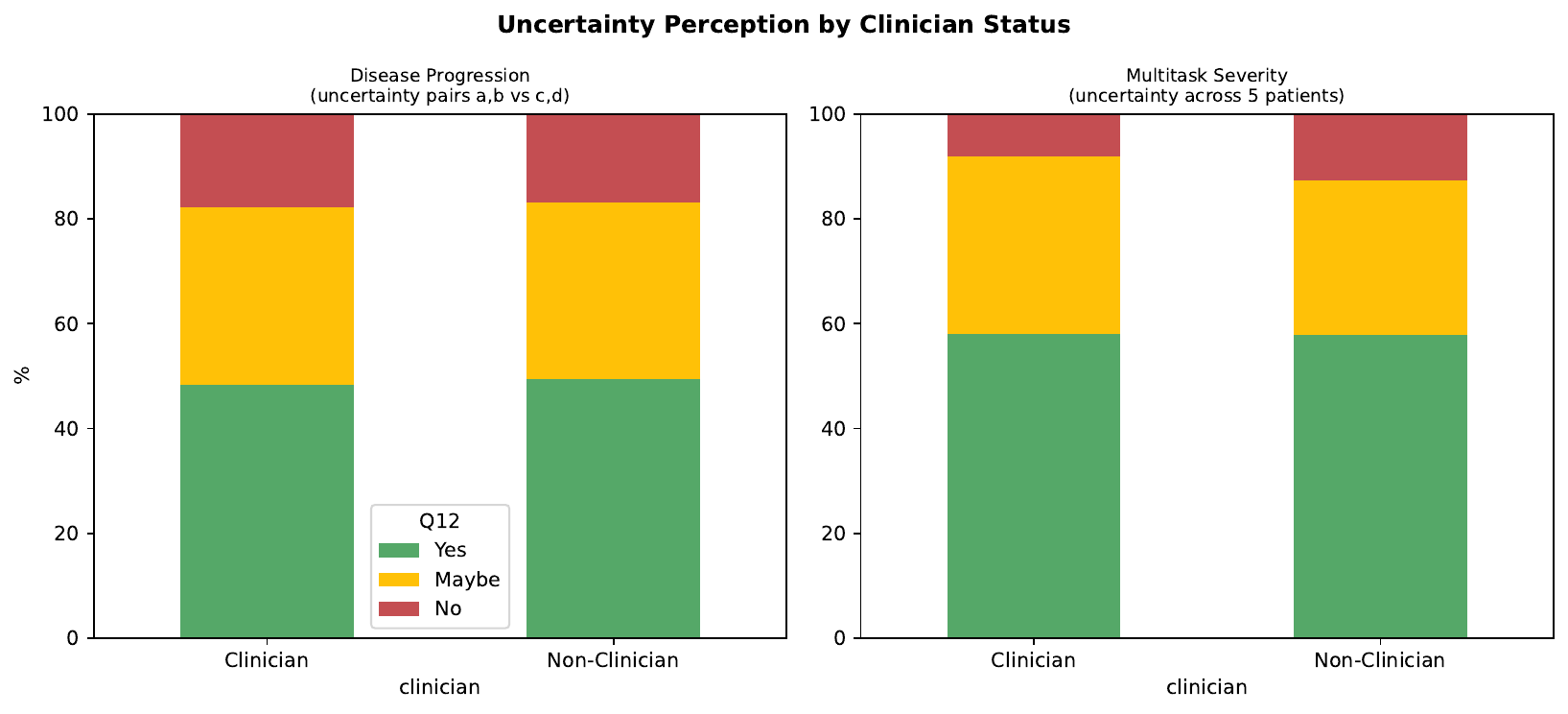}
  \caption{Uncertainty perception (Yes / Maybe / No) stratified by
    clinical background (Clinician vs.\ Non-Clinician) for the
    disease-progression task (left) and multitask scenario (right).}
  \label{fig:uncertainty}
\end{figure}

In the disease-progression task, 49.0\% of respondents reported
definitively observing uncertainty differences between image pairs,
with a further 33.8\% responding ``Maybe'' (total: 82.8\%).
In the multitask scenario, 58.0\% said Yes and 31.2\% said Maybe
(total: 89.2\%). Figure~\ref{fig:uncertainty} shows these distributions
stratified by clinical background. Spearman correlations between uncertainty perception and visual
preference were weak and non-significant: $\rho = 0.05$, $p = 0.563$
(disease progression) and $\rho = 0.15$, $p = 0.065$ (multitask).
Chi-square tests showed no significant differences in uncertainty
perception between clinicians and non-clinicians (both $p_{\text{Bonf}} > 0.05$).

The majority of participants across both clinical and non-clinical groups
perceived uncertainty differences in the VL4ML outputs, confirming
that the black reserved-uncertainty (RU) area and the deviations in prediction
visualizations are effective communication tools for a general audience.
However, uncertainty perception does not predict visual preference,
suggesting that users appreciate the visual format independently of
whether they notice the uncertainty signal specifically. Future work
could make the uncertainty encoding more salient and test whether
stronger uncertainty signals drive larger differences in preference.

\subsection{Logistic Regression: Predictors of Visual Preference}
\label{sec:logreg}

Understanding \emph{who} is most likely to prefer visual AI outputs
allows the system to be personalised. A logistic regression integrates
multiple predictors simultaneously, controlling for confounders and
identifying the independent contribution of each factor. This moves
beyond pairwise comparisons to a multivariate picture of visual
preference.

A binary logistic regression was fitted with visual preference in the
disease-progression task (Q9: Visual = 1, Both/Numeric = 0) as the
outcome and five predictors: clinical background, age, gender,
ease-of-determination (Q8), and uncertainty perception (Q12).
The model converged successfully ($n = 157$). Results are shown in
Table~\ref{tab:logreg}.

\begin{table}[]
\centering
\caption{Logistic regression predicting visual format preference
  (Q9: Visual = 1 vs.\ Other = 0). McFadden pseudo-$R^2 = 0.037$.
  Significant predictor highlighted in bold.}
\label{tab:logreg}
\renewcommand{\arraystretch}{1.2}
\scriptsize
\begin{tabular}{lccccc}
\toprule
\textbf{Predictor} & \textbf{Coef.} & \textbf{SE} & $z$ & $p$ & 95\% CI \\
\midrule
Intercept       &  0.437 & 0.863 &  0.506 & 0.613 & [$-$1.255, 2.129] \\
Clinician       &  0.434 & 0.340 &  1.277 & 0.202 & [$-$0.233, 1.101] \\
\textbf{Age (numeric)} & \textbf{$-$0.377} & \textbf{0.177} & \textbf{$-$2.127} & \textbf{0.033} & [$-$0.724, $-$0.030] \\
Male            &  0.112 & 0.334 &  0.335 & 0.737 & [$-$0.543, 0.767] \\
Ease (Q8)       &  0.210 & 0.223 &  0.942 & 0.346 & [$-$0.227, 0.647] \\
Uncert (Q12)    &  0.019 & 0.228 &  0.085 & 0.932 & [$-$0.427, 0.466] \\
\bottomrule
\multicolumn{6}{l}{LLR $p = 0.150$; AIC = 220.46; $n = 157$}
\end{tabular}
\end{table}

Age is the only statistically significant predictor of visual preference
($\beta = -0.377$, $p = 0.033$): younger participants were more likely
to prefer the visual format. Clinical background, gender, ease of
interpretation, and uncertainty perception did not independently predict
preference once other variables were controlled. The overall model fit
is modest (pseudo-$R^2 = 0.037$, LLR $p = 0.150$), indicating that
the predictors tested here explain only a small portion of the variance
in individual preference — which itself is consistent with the finding
that visual preference is nearly universal across subgroups. Future
models could include technology familiarity, health literacy, or prior
experience with data visualisation as additional predictors.

%% file: sec_disscussion.tex
In this study, we introduced the concept of VL4ML and through four
distinct experiments encompassing multimodal classification, regression
analysis, and predictions, we demonstrated the feasibility and
versatility of this approach. From the perspective of Explainable
AI~2.0, the proposed approach represents a shift from model-centric
explanations toward perception-aligned outputs. Rather than explaining
model decisions after prediction, VL4ML embeds interpretability
directly into the prediction itself, enabling explanations that are
inherently human-readable, context-aware, and actionable.

\paragraph{User preference and ease of interpretation.}
Our survey, conducted with 158 participants (39.2\% clinical workers,
51.7\% female, 84.2\% aged 25--44), affirmed that the visualized
format aligns strongly with user preferences. Merging definitive
(\textit{Yes}) and partial (\textit{Somehow}) agreement into a
single positive-response criterion, at least 79\% of respondents
reported finding each visual display interpretable across all three
clinical tasks, rising to 96.2\% for memorability of the binary ICU
survival display. For the more complex disease-progression task,
54.1\% of participants preferred the visual format exclusively, while
a further 29.3\% preferred receiving both visual and numeric outputs
together -- a total of 83.4\% endorsing visual output as part of
their preferred experience. These figures are consistent with the
picture-superiority effect and with
evidence that visualization accelerates clinical decision-making
by leveraging the brain's preference for perceptual over propositional
representations.

The strong endorsement of the visual format for memorability (84.0\%)
and decision speed (76.9\%) is particularly relevant for clinical
workflows, where rapid recall of prior assessments and time-pressured
decisions are routine. A Spearman rank correlation of $\rho = 0.47$
($p < 0.001$) between memorability endorsement and visual preference
confirms that these two dimensions are internally coherent: users who
find the outputs memorable are also the ones who prefer the format,
suggesting that the design rationale of encoding clinical information
perceptually is working as intended.

\paragraph{Task complexity and interpretability.}
Ease-of-interpretation differed significantly across tasks ($H=28.40$, $p<0.001$). The binary ICU survival task was easiest to interpret, followed by the multitask severity display, while the longitudinal disease-progression display was hardest. Effect sizes were small-to-moderate, suggesting only a modest reduction in usability as task complexity increased.

\paragraph{Subgroup invariance.}
No significant differences were observed between clinicians and non-clinicians across any survey item after Bonferroni correction (all $V<0.10$, all $p_{\text{Bonf}}=1.000$). Similar results were found across gender. The only significant subgroup effect was a modest age-related difference in memorability of the disease-progression display ($V=0.196$), with younger participants rating it as more memorable. These findings support the broad accessibility of VL4ML outputs.

\paragraph{Preference consistency and personalisation.}
Preferences were moderately consistent across tasks, with 49.4\% of participants selecting the same format for both ICU and disease-progression scenarios. Logistic regression identified age as the only independent predictor of visual preference ($p=0.033$), with younger participants favoring visual outputs.

\paragraph{Uncertainty communication.}
Most participants reported being able to identify uncertainty levels in the visualizations (82.8\% for disease progression and 89.2\% for multitask severity). Perceived uncertainty did not differ between clinicians and non-clinicians, indicating that the reserved-uncertainty area and ensemble variability displays effectively communicate model uncertainty to a broad audience.

\paragraph{Future work.} While this study underscores the capabilities and advantages of VL4ML,
several important questions remain for future work. These include the
exploration of optimal image sizes, colors, shapes, and patterns
tailored to each clinical application; the development of more precise
and objective uncertainty quantification metrics; and the investigation
of adaptive visual encodings that maintain interpretability as task
complexity increases. Robust prospective clinical studies are necessary
to rigorously assess real-world effectiveness and impact on patient
outcomes.

Ease of interpretation decreased modestly but significantly with
increasing task complexity, motivating future research into adaptive visual
encodings for complex multi-class and longitudinal clinical outputs.
Individual format preferences were found to be moderately stable
across tasks, supporting the
feasibility of personalised display formatting in clinical deployments.

% ─────────────────────────────────────────────────────────────────────────────
\subsection{Limitations}

While the experiments provide valuable proof of concept and the survey
results align with user preferences, several limitations warrant
consideration.

\begin{enumerate}

\item \textbf{Evaluation scope and observer subjectivity.}
It is important to clarify that VL4ML is not designed as a
replacement for the underlying machine learning model, and
conventional performance metrics such as classification accuracy,
sensitivity, or specificity apply to the ML backbone -- not to the
visualization layer itself. These metrics were intentionally left out
of scope, as the primary contribution of VL4ML is a new
\textit{output representation paradigm}, not a new prediction
algorithm. The clinical models used in our experiments serve as
vehicles for demonstrating the feasibility of the visual output
format across diverse task types.
 
What does require domain-specific evaluation, and remains an open
challenge, is the \textit{human interpretation} of these visual
outputs. Because the rater or clinician serves as the final observer
in our framework, evaluation inherently involves a degree of
subjectivity that varies with the observer's expertise, strictness of
judgment, and familiarity with the specific disease. The
expert-interpretability study reported in Section~\ref{Evaluation}
addresses this directly and provides an initial quantitative benchmark.
However, more comprehensive evaluation protocols -- tailored to each
task type and accounting for inter-rater variability -- represent a
valuable direction for future work as VL4ML matures toward clinical
deployment.

\item \textbf{Uncertainty definition and quantification.}
A simple deep ensemble technique was employed to initiate uncertainty
visualization. While the survey confirmed that 82.8--89.2\% of
participants perceived uncertainty differences in the outputs, a more
precise and objective definition of uncertainty is required. This
quantification should be adaptable to the specific application and
target image design, as uncertainty can manifest differently across
clinical contexts.

\item \textbf{Survey sample limitations.}
The survey included 158 respondents worldwide (39.2\% clinical
workers). Although statistical subgroup analyses found no significant
differences across clinical background or gender, the age distribution
was skewed toward younger adults (84.2\% aged 25--44), and the 55$+$
age group was represented by only 10 participants. A dedicated study
recruiting older adults is needed before conclusions can be
generalised to elderly patient populations, for whom visual AI
interfaces may be especially important.

\item \textbf{Model optimization.}
As this study primarily serves as a proof of concept, network
performance and hyper-parameters were not optimized, and no specific
adjustments were made to address class imbalances. Some observed errors
may therefore be attributable to suboptimal training rather than
fundamental limitations of the VL4ML paradigm.

\end{enumerate}

%% file: sec_conclusion.tex
In this work, we introduced Visualized Learning for Machine Learning (VL4ML), a human-centered explainability framework designed to bridge the gap between machine learning outputs and human understanding in healthcare. Unlike conventional approaches that rely on technical or post-hoc interpretations, VL4ML generates intuitive visual representations in which predictive information and uncertainty are encoded through perceptually meaningful patterns, enabling direct interpretation and providing a complementary medium for uncertainty communication.
We evaluated the proposed paradigm across multiple clinical tasks,
demonstrating its flexibility in handling classification, regression,
longitudinal prediction, and multimodal data. Through a large-scale
human-centered survey ($n = 158$, 39.2\% clinical workers) and an
expert-based interpretability evaluation, we showed that visual
outputs are strongly preferred by users and capable of preserving
clinically meaningful information. Positive reception of the visual
format exceeded 79\% across all tasks and all question dimensions
when combining definitive and partial agreement responses. Visual
outputs were rated as more memorable by 84.0\% and as enabling faster
decision-making by 76.9\% of participants.
Critically, inferential subgroup analyses using chi-square tests with
Bonferroni correction found \textit{no significant differences} in
preference, ease of interpretation, or uncertainty perception between
clinicians and non-clinicians, or between male and female participants.
This universal accessibility is a key strength of the VL4ML
paradigm and distinguishes it from technical XAI methods that require
statistical literacy to interpret.

These findings suggest that aligning AI outputs with human perception
can significantly improve interpretability, trust, and usability in
real-world clinical settings, while enabling intuitive communication
of per-case uncertainty. Such perception-aligned representations are
essential for advancing trustworthy AI systems and supporting their
safe and effective deployment in Healthcare~4.0. By shifting the
focus from model-centric explanations to user-centered representations,
VL4ML contributes to the emerging paradigm of Explainable AI~2.0,
where explanations are inherently human-interpretable, context-aware,
and actionable.

%% file: appendix.tex
%%%%%%%%%%%%%%%%%%%%%%%%%%%%%%%%%%%%%%%%%%%%%%%%%%%%%%%%%%%%%%%%%%%%%%%%%%%%%%%
%%%%%%%%%%%%%%%%%%%%%%%%%%%%%%%%%%%%%%%%%%%%%%%%%%%%%%%%%%%%%%%%%%%%%%%%%%%%%%%
% APPENDIX
%%%%%%%%%%%%%%%%%%%%%%%%%%%%%%%%%%%%%%%%%%%%%%%%%%%%%%%%%%%%%%%%%%%%%%%%%%%%%%%
%%%%%%%%%%%%%%%%%%%%%%%%%%%%%%%%%%%%%%%%%%%%%%%%%%%%%%%%%%%%%%%%%%%%%%%%%%%%%%%
\newpage

\onecolumn

{\LARGE  \textbf{Appendix}}

\section*{Training Loss Curves}

\begin{figure*}[ht]
\centering
\minipage{0.98\textwidth}
  \includegraphics[width=\linewidth,trim={0cm 0cm 0cm 6.5cm},clip]{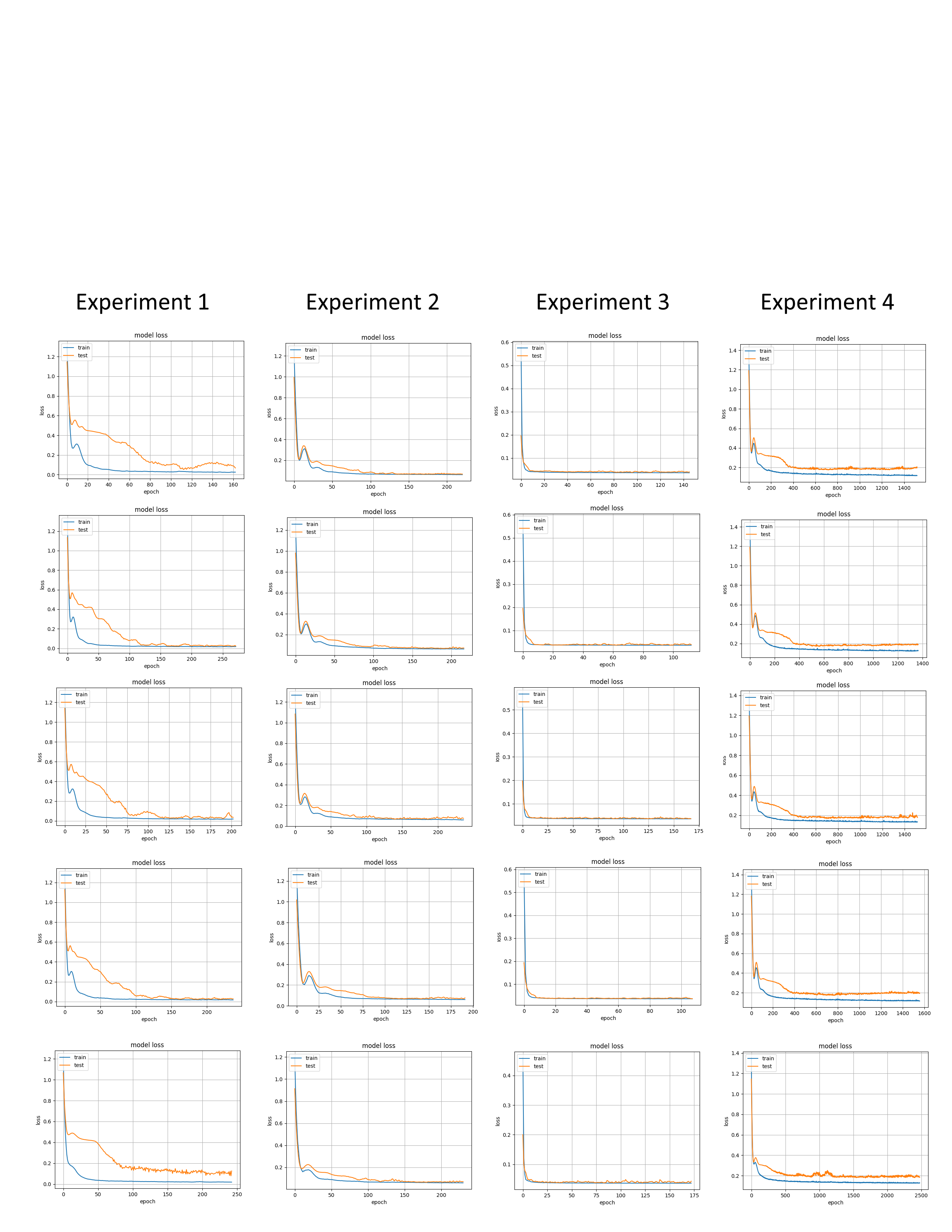}
\endminipage
\caption{The training loss curves for the four experiments are presented. Utilizing the deep ensemble technique, each network undergoes five training iterations with data shuffling.The orange curve represents the loss on the validation set, serving as the test curve.}
\label{fig:results_AD}
\end{figure*}

\newpage

\section*{Details of the experiments and networks}

We have shared all the source codes of the experiments are shared online on GitHub (\href{X}{X}).
For all of the experiments, the network is developed with \textit{Tensorflow} framework. The \textit{Adam} optimizer with a learning rate of 0.001 is used for training and the loss function is the \textit{Mean Absolute Error} between the target image and the produced output. 
In this section, the following acronyms are employed: \textit{FCL} for Forward Convolutional Layers, \textit{TCL} for Transposed Convolutional Layers, and \textit{FC} for Fully Connected Layer.

%%%%%%%%%%%%%%%%%%%%%%%%%%%%%%%%%%%%
\subsection*{Experiment 1 - Patient Survival Prediction in ICU} 

The data used in this example is from \textit{WiDS Datathon 2020} challenge \cite{WiDS, WiDS2} available on \textit{Kaggle} and \textit{PhysioNet} \cite{physionet} website, which focuses on patient health through data from MIT’s GOSSIS (Global Open Source Severity of Illness Score) initiative \cite{GOSSIS}.
Each EHR observation has 186 columns including 3 identification information (\textit{patient id}, \textit{hospital id}, \textit{encounter id}), 16 demographic information  (e.g. \textit{icu type}, \textit{age}, \textit{bmi}, \textit{ethnicity}, \textit{hospital death}), 28 \textit{APACHE} covariates (e.g. \textit{apache 2 diagnosis}, \textit{apache 3j diagnosis}, \textit{gcs eyes apache}, \textit{gcs motor apache}, \textit{resprate apache}, \textit{sodium apache}), 52 vital signs (e.g. \textit{d1 heartrate min}, \textit{d1 resprate max}, \textit{h1 mbp max}, \textit{d1 hemoglobin max} ),  76 lab tests (e.g. \textit{h1 pao2fio2ratio max}, \textit{h1 hemoglobin min}, \textit{h1 glucose min}) and 12 APACHE predictions and comorbidity (\textit{apache 4a hospital death prob}, \textit{cirrhosis}). \\
\textbf{Inputs:} All the elements except identification information and \textit{hospital death} are used as features that create feature vectors with 183 values.  \\
\textbf{Outputs:} As labels, we use information about \textit{hospital deaths}. An image with size 23x23x3 (RGB colorful) is designed for visualization in which white color shows 0 label (survival) and Orange color shows 1 label (fatal). Also, the surrounding pixels are reserved as RU areas with zero intensities (black color) to mean no effect. \\
\textbf{Network:} The network architecture is shown in Figure 2A, including fully connected layers, reshaping, concatenation, and transposed convolution layers with different dilation rates and stride values. Dropout and batch normalization are also used to prevent overfitting.  \\
\textbf{Training:} 15830 observation samples were used and train/test sets were selected randomly by a 66/33 ratio. 10\% of the training set was used as the validation set. The network with 183,336 trainable parameters is trained with 500 maximum epochs and a batch size of 1000. Early stopping regarding the value of training loss and validation loss are also addressed with 30 epochs and 50 epochs of patience, respectively. In the context of deep ensemble training, the network underwent five training iterations with data shuffling.

%%%%%%%%%%%%%%%%%%%%%%%%%%%%%%%%%%%%
\subsection*{Experiment 2- Diabetic Retinopathy Severeness Assessment}

Data is obtained from \textit{DeepDRiD} challenge dataset \cite{Deepdrid}.
The associated dataset contains retinal fundus images and clinicians have rated each image for the severity of diabetic retinopathy on a scale from 0 to 4: 0-No DR, 1-Mild,2 - Moderate,3 - Severe,4 - Proliferative DR. It also contains labels for the quality of the image.
 \\
\textbf{Inputs:} All the \textit{Fundus} images are reshaped into 256x256 resolutions with colors preserved.  \\
\textbf{Outputs:} The desired output is an image with size 23x23x3, and the shape of the output image is an inner square with a color coding for severity levels.  
Five different colors of the inner square are coded to show five severity levels. As shown in Figure \ref{fig:results_retina}, the color of the inner square is bright green (RGB(0,1,0)), dark green (RGB(0.5,1,0)), yellow (RGB(1,1,0)), orange(RGB(1,0.5,0)) and red(RGB(1,0,0)) for 0 to 5 level of disease. To add the quality assessment task, the design includes an additional rectangle whose intensity represents the quality of the Fundus image. As shown in figure \ref{fig:results_retina}A, bright white and dark gray rectangles show high-quality and low-quality images, respectively. 
The rest of the pixels are reserved for \textit{RU} area using black color expressing 0 effect. Two different kinds of outputs are considered for this experiment.\\
\textbf{Network:} The network architecture is the same as shown in Figure 2A for layers L4 and onward. The layers before L4 are as follows: 256x256x3 $\xrightarrow[]{\text{2D-FCL}}$ 127x127x100 $\xrightarrow[]{\text{2D-FCL}}$ 63x63x50 $\xrightarrow[]{\text{2D-FCL}}$ 31x31x50  $\xrightarrow[]{\text{flatten}}$ $\xrightarrow[]{\text{FC}}$ 50 $\xrightarrow[]{\text{FC}}$ 50 $\xrightarrow[]{\text{FC}}$ 100 (L4)$\xrightarrow[]{\text{...}}$\\
\textbf{Training:}  Only the training subset of the dataset. 800 retina Fundus images of 400 subjects are used and the train/test sets are selected randomly by a 66/33 ratio. 10\% of the training set was used as the validation set. The network with 2,511,359 trainable parameters is trained for 500 maximum epochs with  a batch size of 100. 
Early stopping regarding the value of training loss and validation loss are also addressed with 30 epochs and 50 epochs of patience, respectively. In the context of deep ensemble training, the network underwent five training iterations with data shuffling.

%%%%%%%%%%%%%%%%%%%%%%%%%%%%%%%%%%%%
\subsection*{Experiment 3- Estimating the Length-of-Stay in Hospital}

The \textit{MIT MIMIC-III} dataset  \cite{mimic} is used for this experiment because it is an openly available dataset on \textit{Physionet} platform \cite{physionet, mimic_physionet} developed by the MIT Lab for computational physiology, comprising de-identified health data associated with ~60,000 critical care patients. It includes demographics, vital signs, laboratory tests, medications, and more \cite{mimic}. LOS values are defined as the number of days between admission and discharge and just LOS observations lower than 45 are considered for this experiment. The implementation code for this experiment is published online and the details are as follows: \\
\textbf{Inputs:} 52 features are selected as inputs which includes diagnosis categories, 17 primary categories of \textit{ICD-9} (e.g., digestive, skin), demographic information (e.g., age, race, religion, marital status), payment method (e.g., self-pay, medicare), hospital admission type (e.g., urgent, newborn, elective) (\url{https://github.com/daniel-codes/hospital-los-predictor}).  \\
\textbf{Outputs:} A resulting image is 45x45x3 in which each column represents a day (i.e. 45 days) and the column direction is assumed to be the time direction. If the LOS value is $M$, all columns from left to $M^{th}$ column would be cyan color (RGB(0,1,1)). The 10 rows on top and bottom parts of the image are reserved for the \textit{RU} area using black color.  \\
\textbf{Network:} The network architecture is the modified version of the network in 2A as follows. 52 $\xrightarrow[]{\text{FC}}$ 104 $\xrightarrow[]{\text{FC}}$ 52 $\xrightarrow[]{\text{FC}}$ 100  $\xrightarrow[]{\text{Reshape - \dots - L8}}$ 21x21x100 $\xrightarrow[]{\text{L9: 2D-TCL}}$ 43x43x100 $\xrightarrow[]{\text{L10: 2D-TCL}}$ 45x45x3\\
\textbf{Training:} 51944 samples are considered and the train/test sets are selected randomly by a 90/10 ratio. 10\% of the training set was used as the validation set. The network has 137,003 trainable parameters and is trained for 300 maximum epochs with a 300 batch size. Early stopping regarding the value of training loss and validation loss are also addressed with 30 epochs and 50 epochs of patience, respectively.  In the context of deep ensemble training, the network underwent five training iterations with data shuffling.

%%%%%%%%%%%%%%%%%%%%%%%%%%%%%%%%%%%%
\subsection*{Experiment 4- Alzheimer's disease progression}

The clinical data used in this experiment is obtained from the \textit{Alzheimer's Disease Neuroimaging Initiative} (\textit{ADNI}) database (\href{http://adni.loni.usc.edu/}{http://ida.loni.usc.edu}).
Only the subjects that have a baseline scan and showed up for follow-up visits at 6, 12 and 24 months later, have been considered in this longitudinal study.
A total number of 1043 subjects met the follow-up condition, and were categorized into 3 classes AD, MCI, and CN, in baseline and referral sessions.

\textbf{Inputs:} Multi-modal tabular data contains features extracted from \textit{MRI} and \textit{PET} sequences, \textit{demographic information} and some \textit{cognitive measurements} same as reference  \cite{neuroimage_sola, frontiers_sola}. 
\underline{MRI:} Ventricular volume, Hippocampus volume, Whole Brain volume, Entorhinal Cortical thickness, Fusiform, Middle temporal gyrus, and intracranial volume (ICV). 
\underline{PET:} FDG, Pittsburgh Compound-B (PIB), AV45. 
\underline{Cognitive Tests:} Rey Auditory Verbal Learning Test (RAVLT Immediate, RAVLT Learning, RAVLT Forgetting, RAVLT Perc Forgetting), Functional Activities Questionnaires (FAQ), Everyday Cognition (Ecog) scales: (EcogPtMem, EcogPtLang, EcogPtVisspat, EcogPtPlan, EcogPtOrgan, EcogPtDivatt, EcogPtTotal, EcogSPMem, EcogSPLang, EcogSPVisspat, EcogSPPlan, EcogSPOrgan, EcogSPDivatt, and EcogSPTotal). 
\underline{CSF:} Amyloid Beta (ABETA), Phosphorylated Tau Protein (PTAU), and Total Tau Protein (TAU). 
\underline{Risk factors:} Age, years of education, and APOE4.\\
\textbf{Outputs:} A network should automatically create an image containing colorful strips of longitudinal AD assessments where the state of the disease is displayed with different colors, AD: red, MCI: blue, CN: green.
Trajectories of cognitive statues for a 24 month (including baseline T0 and three referral sessions (T1 ($6^{th}$ month), T2 ($12^{th}$ month), T3 ($24^{th}$ month) ) is shown by a sequence of 4 strips in which each strip denotes the disease state and colored as mentioned before. Moreover, the black strip will be appended to the mentioned stripes in order to emphasize the area of \textit{RU} area.  \\
\textbf{Network:} The architecture of the deep learning network has multi-modal tabular inputs and generates a color-coded output image. The size of the output image is 23x23x3 in which the latter 3 channels indicates the \textit{RGB} channels. TThe network is designed so that initial layers concentrate on extracting complex intra-modality features utilizing fully connected layers (FC)(i.e. MRI: 7 $\xrightarrow[]{\text{FC}}$ 14 $\xrightarrow[]{\text{FC}}$ 7  $\xrightarrow[]{\text{}}$ (L3:MRI), PET: 3 $\xrightarrow[]{\text{FC}}$ 6 $\xrightarrow[]{\text{FC}}$ 3 $\xrightarrow[]{\text{}}$ (L3:PET), CSF: 3 $\xrightarrow[]{\text{FC}}$ 6 $\xrightarrow[]{\text{FC}}$ 3 $\xrightarrow[]{\text{}}$ (L3:CSF), COG: PET: 20 $\xrightarrow[]{\text{FC}}$ 40 $\xrightarrow[]{\text{FC}}$ 20 $\xrightarrow[]{\text{}}$ (L3:COG), RF: 3 $\xrightarrow[]{\text{FC}}$ 6 $\xrightarrow[]{\text{FC}}$ 3 $\xrightarrow[]{\text{}}$ (L3:RF) ). Then, feature fusion and feature extraction for the inter-modality phase would be addressed via concatenation and another fully connected layer (concatenation of all L3 layers   $\xrightarrow[]{\text{FC}}$ 100 (L4)). The rest layers are the same as the network in Experiment 1 and Figure 2A. 
\\
\textbf{Training:} The train/test sets are selected randomly by a 90/10 ratio. 10\% of the training set was used as the validation set. The network has 36,143 trainable parameters and is trained for 4000 epochs with a batch size of 500. Early stopping regarding the value of training loss and validation loss are also addressed with 500 epochs and 800 epochs of patience, respectively. In the context of deep ensemble training, the network underwent five training iterations with data shuffling.

%% file: refs.bib
@article{DARPA_XAI,
  title={DARPA’s explainable artificial intelligence (XAI) program},
  author={Gunning, David and Aha, David},
  journal={AI magazine},
  volume={40},
  number={2},
  pages={44--58},
  year={2019}
}

@article{XAI_psychology,
  title={Explainable Artificial Intelligence (XAI): Concepts, taxonomies, opportunities and challenges toward responsible AI},
  author={Arrieta, Alejandro Barredo and D{\'\i}az-Rodr{\'\i}guez, Natalia and Del Ser, Javier and Bennetot, Adrien and Tabik, Siham and Barbado, Alberto and Garc{\'\i}a, Salvador and Gil-L{\'o}pez, Sergio and Molina, Daniel and Benjamins, Richard and others},
  journal={Information fusion},
  volume={58},
  pages={82--115},
  year={2020},
  publisher={Elsevier}
}

@inproceedings{XAI_psychology_2022,
  title={A psychological theory of explainability},
  author={Yang, Scott Cheng-Hsin and Folke, Nils Erik Tomas and Shafto, Patrick},
  booktitle={International Conference on Machine Learning},
  pages={25007--25021},
  year={2022},
  organization={PMLR}
}

@article{TNNLS-2,
  title={Deep Clustering and Visualization for End-to-End High-Dimensional Data Analysis},
  author={Wu, Lirong and Yuan, Lifan and Zhao, Guojiang and Lin, Haitao and Li, Stan Z},
  journal={IEEE Transactions on Neural Networks and Learning Systems},
  year={2022},
  publisher={IEEE}
}

@misc{Deepdrid,
  title={DeepDRiD (Deep Diabetic Retinopathy) challenge},
  publisher={GitHub},
  howpublished = {\url{https://isbi.deepdr.org/data.html}, \url{https://github.com/deepdrdoc/Deep-Diabetic-Retinopathy-Image-Dataset-DeepDRiD-}},
  note = {Accessed: 2020-06-30}
}

@article{neuroimage_sola,
  title={A distributed multitask multimodal approach for the prediction of Alzheimer’s disease in a longitudinal study},
  author={Tabarestani, Solale and Aghili, Maryamossadat and Eslami, Mohammad and Cabrerizo, Mercedes and Barreto, Armando and Rishe, Naphtali and Curiel, Rosie E and Loewenstein, David and Duara, Ranjan and Adjouadi, Malek},
  journal={NeuroImage},
  volume={206},
  pages={116317},
  year={2020},
  publisher={Elsevier}
}

@ARTICLE{frontiers_sola,
AUTHOR={Tabarestani, Solale and Eslami, Mohammad and Cabrerizo, Mercedes and Curiel, Rosie E. and Barreto, Armando and Rishe, Naphtali and Vaillancourt, David and DeKosky, Steven T. and Loewenstein, David A. and Duara, Ranjan and Adjouadi, Malek},   
TITLE={A Tensorized Multitask Deep Learning Network for Progression Prediction of Alzheimer’s Disease},      
JOURNAL={Frontiers in Aging Neuroscience},      
VOLUME={14},      
YEAR={2022},      
URL={https://www.frontiersin.org/article/10.3389/fnagi.2022.810873},       
DOI={10.3389/fnagi.2022.810873},      
	
ISSN={1663-4365}
}

@article{longo2024explainable,
  title={Explainable Artificial Intelligence (XAI) 2.0: A manifesto of open challenges and interdisciplinary research directions},
  author={Longo, Luca and Brcic, Mario and Cabitza, Federico and Choi, Jaesik and Confalonieri, Roberto and Del Ser, Javier and Guidotti, Riccardo and Hayashi, Yoichi and Herrera, Francisco and Holzinger, Andreas and others},
  journal={Information Fusion},
  volume={106},
  pages={102301},
  year={2024},
  publisher={Elsevier}
}

@article{panayides2025position,
  title={Position paper: Artificial intelligence in medical image analysis: Advances, clinical translation, and emerging frontiers},
  author={Panayides, AS and Chen, H and Filipovic, ND and Geroski, T and Hou, J and Lekadir, K and Marias, K and Matsopoulos, GK and Papanastasiou, Giorgos and Sarder, P and others},
  journal={IEEE Journal of Biomedical and Health Informatics},
  year={2025},
  publisher={IEEE}
}

@article{ghanvatkar2024evaluating,
  title={Evaluating explanations from AI algorithms for clinical decision-making: a social science-based approach},
  author={Ghanvatkar, Suparna and Rajan, Vaibhav},
  journal={IEEE Journal of Biomedical and Health Informatics},
  volume={28},
  number={7},
  pages={4269--4280},
  year={2024},
  publisher={IEEE}
}

@article{dwivedi2023explainable,
  title={Explainable AI (XAI): Core ideas, techniques, and solutions},
  author={Dwivedi, Rudresh and Dave, Devam and Naik, Het and Singhal, Smiti and Omer, Rana and Patel, Pankesh and Qian, Bin and Wen, Zhenyu and Shah, Tejal and Morgan, Graham and others},
  journal={ACM computing surveys},
  volume={55},
  number={9},
  pages={1--33},
  year={2023},
  publisher={ACM New York, NY}
}

@article{huang2024review,
  title={A review of uncertainty quantification in medical image analysis: Probabilistic and non-probabilistic methods},
  author={Huang, Ling and Ruan, Su and Xing, Yucheng and Feng, Mengling},
  journal={Medical Image Analysis},
  volume={97},
  pages={103223},
  year={2024},
  publisher={Elsevier}
}

@article{wang2023calibration,
  title={Calibration in deep learning: A survey of the state-of-the-art},
  author={Wang, Cheng},
  journal={arXiv preprint arXiv:2308.01222},
  year={2023}
}

@article{he2026survey,
  title={A survey on uncertainty quantification methods for deep learning},
  author={He, Wenchong and Jiang, Zhe and Xiao, Tingsong and Xu, Zelin and Li, Yukun},
  journal={ACM Computing Surveys},
  volume={58},
  number={7},
  pages={1--35},
  year={2026},
  publisher={ACM New York, NY}
}

@article{salvi2025explainability,
  title={Explainability and uncertainty: Two sides of the same coin for enhancing the interpretability of deep learning models in healthcare},
  author={Salvi, Massimo and Seoni, Silvia and Campagner, Andrea and Gertych, Arkadiusz and Acharya, U Rajendra and Molinari, Filippo and Cabitza, Federico},
  journal={International Journal of Medical Informatics},
  volume={197},
  pages={105846},
  year={2025},
  publisher={Elsevier}
}

@inproceedings{bibal2022attention,
  title={Is attention explanation? an introduction to the debate},
  author={Bibal, Adrien and Cardon, R{\'e}mi and Alfter, David and Wilkens, Rodrigo and Wang, Xiaoou and Fran{\c{c}}ois, Thomas and Watrin, Patrick},
  booktitle={Proceedings of the 60th Annual Meeting of the Association for Computational Linguistics (volume 1: long papers)},
  pages={3889--3900},
  year={2022}
}

@inproceedings{koh2020concept,
  title={Concept bottleneck models},
  author={Koh, Pang Wei and Nguyen, Thao and Tang, Yew Siang and Mussmann, Stephen and Pierson, Emma and Kim, Been and Liang, Percy},
  booktitle={International conference on machine learning},
  pages={5338--5348},
  year={2020},
  organization={PMLR}
}

@book{molnar2020interpretable,
  title={Interpretable machine learning},
  author={Molnar, Christoph},
  year={2020},
  publisher={Lulu. com}
}

@article{eslami2023unique,
  title={A unique color-coded visualization system with multimodal information fusion and deep learning in a longitudinal study of Alzheimer's disease},
  author={Eslami, Mohammad and Tabarestani, Solale and Adjouadi, Malek},
  journal={Artificial intelligence in medicine},
  volume={140},
  pages={102543},
  year={2023},
  publisher={Elsevier}
}

@article{bias,
  title={Artificial intelligence, bias and clinical safety},
  author={Challen, Robert and Denny, Joshua and Pitt, Martin and Gompels, Luke and Edwards, Tom and Tsaneva-Atanasova, Krasimira},
  journal={BMJ Qual Saf},
  volume={28},
  number={3},
  pages={231--237},
  year={2019},
  publisher={BMJ Publishing Group Ltd}
}

@article{mimic,
  title={MIMIC-III, a freely accessible critical care database},
  author={Johnson, Alistair EW and Pollard, Tom J and Shen, Lu and Li-wei, H Lehman and Feng, Mengling and Ghassemi, Mohammad and Moody, Benjamin and Szolovits, Peter and Celi, Leo Anthony and Mark, Roger G},
  journal={Scientific data},
  volume={3},
  pages={160035},
  year={2016},
  publisher={Nature Publishing Group}
}

@article{physionet,
  title={PhysioBank, PhysioToolkit, and PhysioNet: components of a new research resource for complex physiologic signals},
  author={Goldberger, Ary L and Amaral, Luis AN and Glass, Leon and Hausdorff, Jeffrey M and Ivanov, Plamen Ch and Mark, Roger G and Mietus, Joseph E and Moody, George B and Peng, Chung-Kang and Stanley, H Eugene},
  journal={circulation},
  volume={101},
  number={23},
  pages={e215--e220},
  year={2000},
  publisher={Am Heart Assoc}
}

@article{patient_pref_ieee_review,
  title={Data-driven healthcare: Challenges and opportunities for interactive visualization},
  author={Gotz, David and Borland, David},
  journal={IEEE computer graphics and applications},
  volume={36},
  number={3},
  pages={90--96},
  year={2016},
  publisher={IEEE}
}

@article{patient_pref_2,
  title={Patient preferences for visualization of longitudinal patient-reported outcomes data},
  author={Stonbraker, Samantha and Porras, Tiffany and Schnall, Rebecca},
  journal={Journal of the American Medical Informatics Association},
  volume={27},
  number={2},
  pages={212--224},
  year={2020},
  publisher={Oxford University Press}
}

@article{patients_pref_kid_3,
  title={Readability and visuals in medical research information forms for children and adolescents},
  author={Grootens-Wiegers, Petronella and De Vries, Martine C and Vossen, Tessa E and Van den Broek, Jos M},
  journal={Science Communication},
  volume={37},
  number={1},
  pages={89--117},
  year={2015},
  publisher={Sage Publications Sage CA: Los Angeles, CA}
}

@article{Nature_UncertaintyQuanti,
  title={The need for uncertainty quantification in machine-assisted medical decision making},
  author={Begoli, Edmon and Bhattacharya, Tanmoy and Kusnezov, Dimitri},
  journal={Nature Machine Intelligence},
  volume={1},
  number={1},
  pages={20--23},
  year={2019},
  publisher={Nature Publishing Group}
}

@misc{WiDS,
  title={2020 Global Women in Data Science (WiDS) Conference, Datathon Challenge},
  publisher={Kaggle},
  howpublished = {\url{https://www.kaggle.com/c/widsdatathon2020/}},
  note = {Accessed: 2020-06-30}
}

@misc{WiDS2,
  doi = {10.13026/VC0E-TH79},
  url = {https://physionet.org/content/widsdatathon2020/},
  author = {Lee,  Meredith and Raffa,  Jesse and Ghassemi,  Marzyeh and Pollard,  Tom and Kalanidhi,  Sharada and Badawi,  Omar and Matthys,  Karen and Celi,  Leo Anthony},
  title = {WiDS (Women in Data Science) Datathon 2020: ICU Mortality Prediction},
  publisher = {physionet.org},
  year = {2020}
}

@inproceedings{defer2,
  title={Predict Responsibly: Improving Fairness and Accuracy by Learning to Defer},
  author={David Madras and T. Pitassi and R. Zemel},
  booktitle={NeurIPS},
  year={2018}
}

@article{defer3,
  title={Second opinion needed: communicating uncertainty in medical machine learning},
  author={Kompa, Benjamin and Snoek, Jasper and Beam, Andrew L},
  journal={NPJ Digital Medicine},
  volume={4},
  number={1},
  pages={1--6},
  year={2021},
  publisher={Nature Publishing Group}
}

@misc{mimic_physionet,
  title={mimic III on physionet},
  publisher={physionet},
  howpublished = {\url{https://physionet.org/content/mimiciii-demo/1.4/} \& \url{https://mimic.physionet.org/}},
  note = {Accessed: 2020-06-30}
}

@article{GOSSIS,
  title={33: THE GLOBAL OPEN SOURCE SEVERITY OF ILLNESS SCORE (GOSSIS)},
  author={Raffa, Jesse and Johnson, Alistair and Celi, Leo Anthony and Pollard, Tom and Pilcher, David and Badawi, Omar},
  journal={Critical Care Medicine},
  volume={47},
  number={1},
  pages={17},
  year={2019},
  publisher={LWW},
  howpublished = {\url{https://gossis.mit.edu/}},
}

@article{baadte2019picture,
  title={The picture superiority effect in associative memory: A developmental study},
  author={Baadte, Christiane and Meinhardt-Injac, Bozana},
  journal={British Journal of Developmental Psychology},
  year={2019},
  publisher={Wiley Online Library}
}

@article{grady1998neural,
  title={Neural correlates of the episodic encoding of pictures and words},
  author={Grady, Cheryl L and McIntosh, Anthony R and Rajah, M Natasha and Craik, Fergus IM},
  journal={Proceedings of the National Academy of Sciences},
  volume={95},
  number={5},
  pages={2703--2708},
  year={1998},
  publisher={National Acad Sciences}
}

@article{potter2014detecting,
  title={Detecting meaning in RSVP at 13 ms per picture},
  author={Potter, Mary C and Wyble, Brad and Hagmann, Carl Erick and McCourt, Emily S},
  journal={Attention, Perception, \& Psychophysics},
  volume={76},
  number={2},
  pages={270--279},
  year={2014},
  publisher={Springer}
}

@book{lindsay2013human-book,
  title={Human information processing: An introduction to psychology},
  author={Lindsay, Peter H and Norman, Donald A},
  year={2013},
  publisher={Academic press}
}

@article{padilla2018decision-review,
  title={Decision making with visualizations: a cognitive framework across disciplines},
  author={Padilla, Lace M and Creem-Regehr, Sarah H and Hegarty, Mary and Stefanucci, Jeanine K},
  journal={Cognitive research: principles and implications},
  volume={3},
  number={1},
  pages={29},
  year={2018},
  publisher={SpringerOpen}
}

@article{cardoso2016graph,
  title={A graph is worth a thousand words: How overconfidence and graphical disclosure of numerical information influence financial analysts accuracy on decision making},
  author={Cardoso, Ricardo Lopes and Leite, Rodrigo Oliveira and de Aquino, Andr{\'e} Carlos Busanelli},
  journal={PloS one},
  volume={11},
  number={8},
  pages={e0160443},
  year={2016},
  publisher={Public Library of Science}
}

@article{IEEE_MDvisualization,
  title={Ten open challenges in medical visualization},
  author={Gillmann, Christina and Smit, Noeska N and Gr{\"o}ller, Eduard and Preim, Bernhard and Vilanova, Anna and Wischgoll, Thomas},
  journal={IEEE Computer Graphics and Applications},
  volume={41},
  number={5},
  pages={7--15},
  year={2021},
  publisher={IEEE}
}

@article{deep_ensembles,
  title={Uncertainty quantification and deep ensembles},
  author={Rahaman, Rahul and others},
  journal={Advances in Neural Information Processing Systems},
  volume={34},
  pages={20063--20075},
  year={2021}
}
